\UseRawInputEncoding
\documentclass[11pt]{article}

\usepackage[final]{acl}
\usepackage[most]{tcolorbox}

\usepackage{times}
\usepackage{latexsym}
\usepackage{enumitem}
\usepackage{needspace}
\usepackage{algorithm}
\usepackage{algpseudocode}
\usepackage{amssymb}

\usepackage[T1]{fontenc}

\usepackage{hyperref}

\usepackage{microtype}

\usepackage{graphicx}
\usepackage{hyperref}
\usepackage{url}
\usepackage{booktabs}
\usepackage{multirow}
\usepackage{adjustbox}

\tcbset{
    listing options={
        basicstyle=\fontfamily{qtm}\selectfont\small\ttfamily,
        breaklines=true
    },
    fontupper=\fontfamily{qtm}\selectfont
}

\title{Programming over Thinking: Efficient and Robust \\Multi-Constraint Planning}

\author{
Derrick Goh Xin Deik$^{1}$, Quanyu Long$^{1}$, 
Zhengyuan Liu$^{2}$, Nancy F. Chen$^{2}$, Wenya Wang$^{1}$\thanks{Corresponding author} \\
$^{1}$Nanyang Technological University, Singapore \\
$^{2}$Agency for Science, Technology and Research (A*STAR), Singapore \\
\texttt{gohx0043@e.ntu.edu.sg, wangwy@ntu.edu.sg}
}

\begin{document}
\maketitle
\begin{abstract}
Multi-constraint planning involves identifying, evaluating, and refining candidate plans while satisfying multiple, potentially conflicting constraints. Existing large language model (LLM) approaches face fundamental limitations in this domain. Pure reasoning paradigms, which rely on long natural language chains, are prone to inconsistency, error accumulation, and prohibitive cost as constraints compound. Conversely, LLMs combined with coding- or solver-based strategies lack flexibility: they often generate problem-specific code from scratch or depend on fixed solvers, failing to capture generalizable logic across diverse problems. 
To address these challenges, we introduce the \textbf{S}calable \textbf{CO}de \textbf{P}lanning \textbf{E}ngine (SCOPE), a framework that disentangles query-specific reasoning from generic code execution. By separating reasoning from execution, SCOPE produces solver functions that are consistent, deterministic, and reusable across queries while requiring only minimal changes to input parameters. SCOPE achieves state-of-the-art performance while lowering cost and latency. For example, with GPT-4o, it reaches 93.1\% success on TravelPlanner, a 61.6\% gain over the best baseline (CoT) while cutting inference cost by 1.4x and time by ~4.67x. Code is available at \href{https://github.com/DerrickGXD/SCOPE}{https://github.com/DerrickGXD/SCOPE}. 
\end{abstract}

\begin{figure*}[t!]
    \centering
    \setlength{\abovecaptionskip}{0.2cm}
    \setlength{\belowcaptionskip}{-0.3cm}
    \includegraphics[width=1.0\textwidth, height=0.63\textwidth]
    {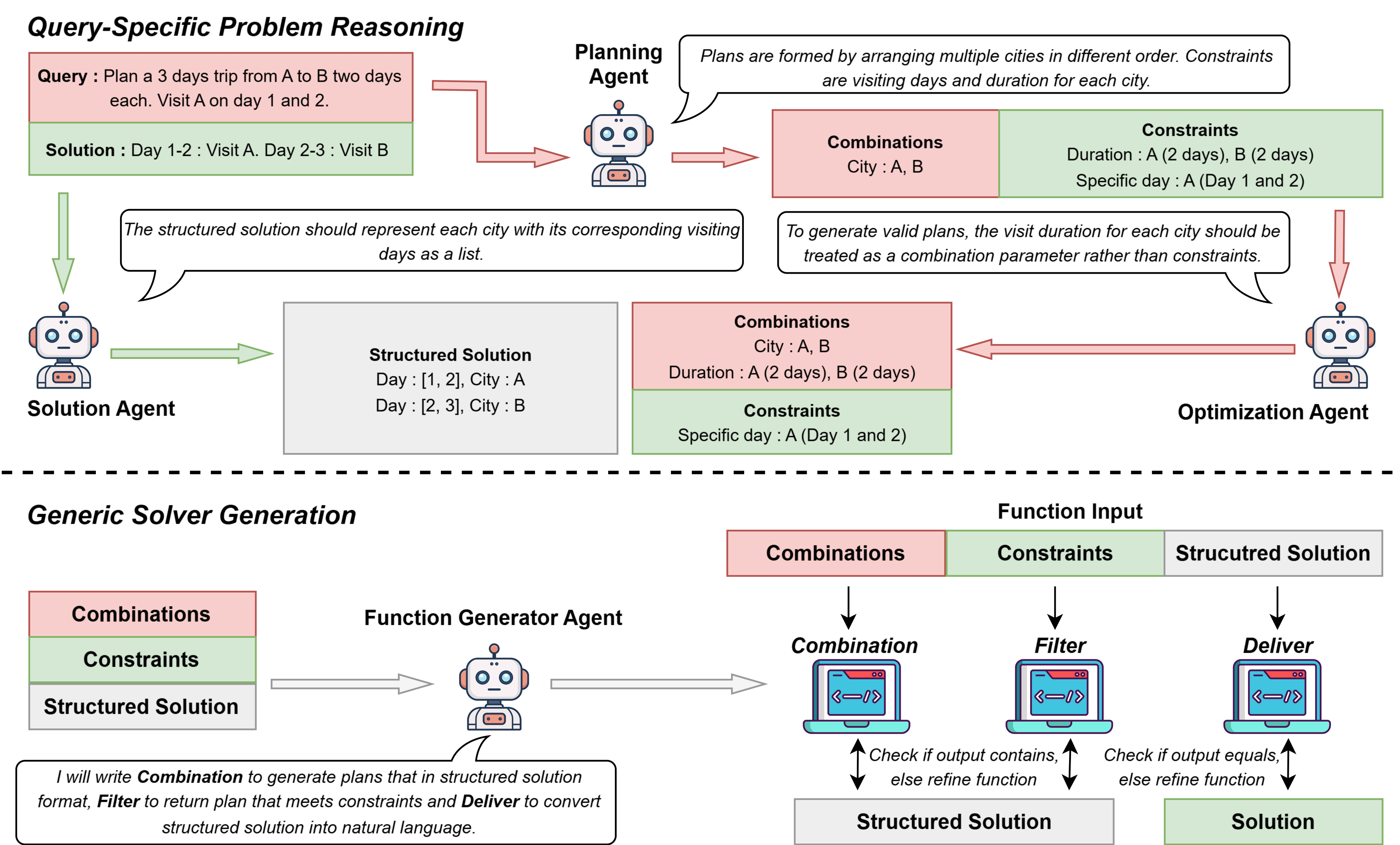}
    \caption{The overview workflow of \textbf{SCOPE}. The workflow consists of 2 stages: \textbf{Query-Specific Problem Reasoning} (top-half) and \textbf{Generic Solver Generation} (bottom-half). Different colors of arrows represent different workflow construction: (red, structured representations), (green, structured solution), (gray, solver functions).}
    \label{fig:scope_diagram}
\vspace{-0.3cm}
\end{figure*}

\section{Introduction}
Planning is the process by which an agent organizes sequences of decisions or actions to achieve a goal. With the rapid advancement of LLMs, there has been growing interest in applying them to automate planning tasks \cite{Wei2025PlanGen}. Unlike traditional problem-solving tasks such as question answering or math reasoning, planning demands exploration of a rapidly expanding solution space, where small errors accumulate and lead to inconsistent or invalid outcomes. Recent development of LLMs in text-based reasoning \cite{Wei2022ChainOfThought, Yao2023ToT, Yuan2025Evoagent, gui2025HTP}, powered by strong reasoning capabilities, has shown promise on simpler planning tasks. However, due to LLMs output being inherently probabilistic, 
these methods lack robustness, where small variations of reasoning paths can produce inconsistent or invalid solutions, and models often lose track of constraints or accumulate errors over long reasoning chains. In addition, they face scalability issues. Reasoning chains grow exponentially with task difficulty, require large numbers of tokens, and become increasingly costly to execute (refer to Figure \ref{fig:all_cost_time}).  

To address these issues, prior work has explored integrating external solvers into the reasoning process \cite{Chen2023PoT,Jiang2024SelfPlanning}, leveraging their reliability and determinism to enforce constraints and verify solutions. Subsequent efforts have aimed to handle diverse constraints in planning \cite{Hao2025Planning,Hao2025Formal,Wang2025PEA}. Existing methods either perform similar reasoning in natural language or generate similar solver code for each query. As a result, they do not support reusable execution abstractions and incur high computational cost. Nevertheless, existing code-based methods lack a systematic mechanism to capture the underlying logic of multi-constraint planning, resulting in inefficiency and increased error.

To this end, we propose the \textbf{S}calable \textbf{CO}de \textbf{P}lanning \textbf{E}ngine (SCOPE), a multi-agent framework that employs a two-stage reasoning process to convert natural language queries into optimized structured representations, which are then processed by reusable solver functions. In the problem reasoning phase, SCOPE utilizes \textbf{Problem Formalization} and \textbf{Problem Optimization} to translate queries into structured representations consisting combination and constraint parameters. Combination parameters specify elements and properties required to generate candidate plans, whereas constraint parameters define conditions these plans must satisfy. LLM agents use these representations to generate general solver functions. Subsequent queries within the same domain are similarly converted into structured representations, enabling subsequent queries in the same domain to be solved efficiently without regenerating code.

To enhance reliability and eliminate dependence on labor-intensive expert prompt design, SCOPE incorporates a parameter-free refinement step in which a single example query and answer are used for output reflection to automatically adjust prompts. This design establishes a fully automated pipeline wherein LLM agents autonomously generate optimized representations, construct reusable solvers, and deliver consistent, scalable solutions to new problems without additional training or manual intervention. Unlike ad-hoc reasoning or problem-specific coding, this approach ensures both flexibility and robustness, as reasoning is tailored to individual queries, while solver functions remain stable across queries in the same domain.

We evaluate SCOPE on benchmarks such as TravelPlanner \cite{Xie2024TravelPlanner} and Natural Plan \cite{Zheng2024NaturalPlan}, and reach state-of-the-art results across 5 recent models with much lower inference cost. 
Our contributions are threefold:

\begin{itemize}[itemsep=2pt, parsep=0pt, topsep=2pt]
  \item We propose a reusable abstraction that separates query-specific reasoning from execution logic. It captures the key parameters that generate or filter candidate plans, enabling solver functions that can handle different queries with minimal adaptation.
  
  \item We implement an autonomous multi-agent pipeline that induces this abstraction from a single example. The pipeline identifies relevant combinations and constraints and produces solver functions that operate on parameter values rather than query-specific content, allowing the abstraction to be applied broadly.
  
  \item We provide empirical evidence that this approach improves robustness, reduces computational cost, and scales efficiently across benchmarks and models.
\end{itemize}

\section{Related Work}
Recent advances in LLMs have sparked growing interest in their reasoning and planning capabilities \cite{achiam2023gpt,dong2024survey}. A prominent line of work studies text-based reasoning through prompting strategies to produce explicit, step-by-step intermediate reasoning \cite{Wei2022ChainOfThought, shinn2023reflexion, Yao2023React, Madaan2023SelfRefine, chen2024toward, lee2025evolving} in natural language. Another related direction investigates multi-agent planning \cite{Chen2024Agentverse, Chen2024AutoAgents, WangSPP2024, cong2024Pmc, gui2025HTP}. These work allow planning to be more manageable by dividing the tasks into components managed by expert agents. 

Text-based reasoning relies on implicit LLM reasoning and does not explicitly enumerate the solution space, making it prone to errors in multi-constraint planning \cite{Kambhampati2024Plan}. Benchmarks \cite{valmeekam2023planbench, Zheng2024NaturalPlan, Xie2024TravelPlanner} show failures under long-horizon reasoning and complex constraints. Multi-agent approaches also require extensive prompt engineering and struggle to systematically decompose such constraints \cite{han2024llm}.

An alternative direction leverages code generation as proxy for reasoning \cite{Chen2023PoT, Yang2025Code2Think, liu2024interactive, Yang2023Coupling}. These works proposes formalizing problem and replacing reasoning with code to minimize errors. Some recent works have attempted to use code or Planning Domain Definition Language (PDDL) to help reasoning of LLM on constraint satisfaction problems \cite{guan2023leveraging, Katz2024Tos, Wang2025PEA, Hao2025Formal, Hao2025Planning}. While these methods improves performance, they rely heavily on manual prompting and lack generalization when applied to planning tasks in new domains.

\section{Methodology}

\begin{figure*}[t]
    \centering
    \setlength{\belowcaptionskip}{-0.3cm}
    \includegraphics[width=1.0\textwidth, height=0.65\textwidth]{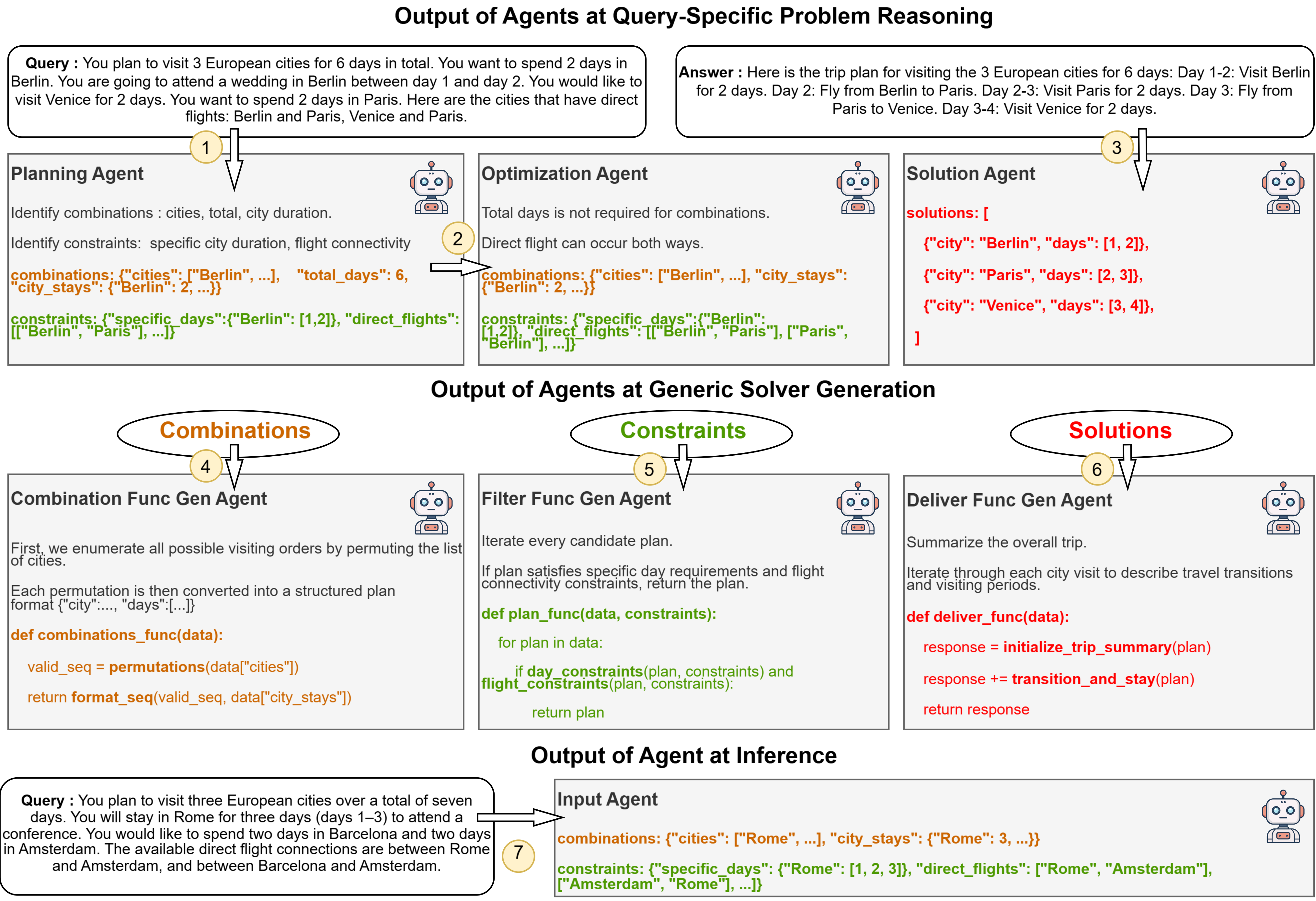}
    \caption{The output of agents at two different stages of \textbf{SCOPE} and inference.}
    
    \label{fig:output_diagram}
\end{figure*}

\subsection{Overview}
\label{sec:method:solver}
Multi-constraint planning requires exhaustive reasoning over a large candidate space, where missing a single combination can lead to incorrect results. Text-based reasoning methods often fail to systematically enumerate all candidates and miss valid plans, while existing solver-based approaches ensure completeness but incur high inference cost by generating solver code per query. In contrast, our framework leverages LLMs to construct reusable solver functions.

We propose a fully autonomous solver generation and refinement framework mechanism that enables the system to independently discover effective instructions without manual intervention. The framework consists of two high-level stages, namely \textbf{Query-Specific Problem Reasoning} and \textbf{Generic Solver Generation}. As shown in Figure \ref{fig:scope_diagram}, the Query-Specific Problem Reasoning stage aims to generate a formal representation of the problem domain, including combinations, constraints and structured solution, enabling the system to autonomously design the solver. 
Given the formal representation, the Generic Solver Generation stage aims to construct the functions required to solve the problem domain, 
namely \emph{Combination Function}, \emph{Filter Function} and \emph{Deliver Function}. 

\subsection{Query-Specific Problem Reasoning}
\label{sec:method:problem_reasoning}

In real-world scenarios, explicit reasoning or structured solver logic is not provided to the model, so the model must learn to infer this from examples in order to produce reusable functions for new queries. This stage utilizes a single example query to extract the key parameters and reasoning needed to construct solver functions for the given problem domain. This stage comprises two sequential steps, \emph{Problem Formalization} and \emph{Problem Optimization}. 

\paragraph{Problem Formalization.} As the first step, Problem Formalization is designed to translate example queries and answers into structured JSON representations that define the input and output interfaces for solver function construction in a later stage. It employs two LLM agents, namely \textbf{Planning Agent} and \textbf{Solution Agent}. Particularly, Planning Agent converts the example query into structured representation with two keys, \texttt{combinations} and \texttt{constraints}.  \texttt{Combinations} contains the parameters for complete candidate generation space, ensuring exhaustive enumeration rather than heuristic exploration. \texttt{Constraints} contains parameters for filtering plans,  isolating validation logic from generation logic to enable deterministic filtering. Considering the example query as shown in Figure \ref{fig:output_diagram}, Planning Agent identifies that candidate plans can be generated by arranging the different cities (Berlin, Venice and Paris) in different visiting orders, and therefore proposes combination parameters consisting of "cities", "city\_stays", etc. Similarly, a plan is only valid under certain constraints, hence Planning Agent proposes constraint parameters such as "specific\_days".

Simultaneously, the \textbf{Solution Agent} receives the example answer and converts it into a structured representation, which defines the expected output format used in solver function construction in the next stage. Considering the Figure \ref{fig:output_diagram} example, the structured representation would be a list of JSON, which the key "city" and "days" outlines the name of city and visiting days. Together, the Planning Agent and Solution Agent determine the input and output formats of all solver functions.

\paragraph{Problem Optimization.} Without optimization, structured representations often contain redundant, entangled, or underspecified parameters that can prevent correct solver function synthesis. SCOPE employs one or more \textbf{Optimization Agents}, each assigned a distinct role via its prompt, to iteratively refine the parameters. These agents remove redundancy, resolve ambiguities, and ensure completeness, guaranteeing that combinations fully cover the candidate plan space while constraints remain clear, consistent, and non-conflicting. The prompts provided to each Optimization Agent are described in Appendix \ref{app:scope_prompt}.



\subsection{Generic Solver Generation}
\label{sec:method:problem_reasoning}

Constructing reusable solver functions is challenging because the reasoning must be abstracted from any specific query. To address this, SCOPE decomposes the solver into distinct functions, each responsible for a single task. This decomposition allows the system to focus supervision on ensuring correctness for each function. This stage proceeds through two sequential phases, \emph{Solver Construction} and \emph{Solver Refinement}. In particular, we generate the following solver functions in this stage:
\vspace{-0.5mm}
\begin{itemize}[itemsep=2pt, parsep=0pt, topsep=2pt]
    \item \textbf{Combination Function}: takes as input \texttt{data}, which is the values of the \texttt{combinations} parameters, and generates candidate plans in structured JSON form.
    \item \textbf{Filter Function}: takes as input \texttt{data}, the candidate plans generated by Combination Function, and \texttt{constraints}, which is the values of the \texttt{constraints} parameters, and returns the plan that satisfy all constraints.
    \item \textbf{Deliver Function}: takes as input \texttt{data}, the plan returned by Filter Function and converts it into a natural language answer that follows the desired output format.
\end{itemize}

\paragraph{Solver Construction.} We employ three agents, \textbf{Combination Function Generator Agent}, \textbf{Filter Function Generator Agent} and \textbf{Deliver Function Generator Agent} to generate Combination, Filter and Deliver Function respectively. The Combination Function Generator Agent is provided the \texttt{combinations} parameters and their descriptions produced by the Planning Agent, along with the structured solution generated by the Solution Agent.

The Combination Function Generator Agent leverages these information to construct Combination Function that produces candidate plans that adhere to the required structure. For instance, in Figure \ref{fig:output_diagram}, the agent constructs the Combination Function to generate all permutations of the "cities" parameter to enumerate possible visiting orders, and then assigns visiting days using "city\_stays". Each resulting plan conforms to the structured solution designed by Solution Agent.

Similarly, the Filter Function Generator Agent receives the \texttt{constraints} parameters and their descriptions, together with the same structured solution. Using this information, it constructs the Filter Function, which extracts relevant attributes from each plan and compares them against the specified constraint values. In Figure \ref{fig:output_diagram}, the generated Filter Function iterates through each plan produced by the Combination Function and checks whether the "specific\_days" and "direct\_flights" constraint requirements are satisfied by each plan, returning only the plan that meet all constraints.

The Deliver Function Generator Agent receives the structured solution along with ground-truth answer, enabling it to generate Deliver Function that converts any structured solution in the same format into a natural-language answer consistent with the desired output.

\paragraph{Solver Refinement.} The initial functions generated by the Combination and Filter Function Generator Agents may be suboptimal. We refine them using the same example query and solution from the previous stage, with the Planning Agent’s structured representation and the Solution Agent’s structured solution as supervision. Refinement is applied when a function fails, defined as follows:
\begin{itemize}[itemsep=2pt, parsep=0pt, topsep=2pt]
    \item \textbf{Combination Function}: The list of candidate plans generated by the function does not contain the structured solution.
    \item \textbf{Filter Function}: The plan returned by function does not match the structured solution.
    \item \textbf{Deliver Function}: The natural language answer produced by the function does not match the ground-truth answer.
\end{itemize}

By comparing the function outputs with the structured solution or ground-truth answer, the agents iteratively regenerate and adjust their functions until the expected output is produced. 



We note that using a single example of query and solution during solver refinement does not cause overfitting. The Function Generator Agent is instructed to design functions that operate on input parameters, rather than hardcoding instance-specific values. Likewise, the Planning Agent constructs \texttt{combinations} and \texttt{constraints} to capture the general domain structure, rather than the specific keys or values of the example. Consequently, the generated solver functions generalize across queries, reflecting structural constraints rather than memorizing a single instance. For applications requiring additional assurance, solver functions can be validated and refined against multiple examples.

In our experiments, a single example suffices as it contains full constraint coverage. To analyze the effect of using multiple examples in SCOPE, we conduct experiments on Trip Planning solver construction using individual examples with incomplete constraint coverage, as well as both examples together. Our analysis in Appendix \ref{app:multi_example_analysis} shows SCOPE can leverage multiple examples to improve robustness via union constraint coverage.

\subsection{Inference}


The query and structured representation pair from Query-Specific Problem Reasoning stage is used as a single in-context exemplar that specifies the expected structured output format for inference. Specifically, at inference, the framework first uses a single LLM as the \textbf{Input Agent}, which takes the exemplar query, its corresponding \texttt{combinations} and \texttt{constraints} optimized from the problem reasoning stage as the one-shot prompt. Along with the test query appended in the prompt, the Input Agent generates \texttt{combinations} and \texttt{constraints} for the test query. These are further fed into the reusable solver functions generated by the SCOPE workflow to produce the final answer. 

During inference, the agent does not generate or modify any solver code. Instead, it applies the pre-generated solver functions to produce the final solution. The inferred \texttt{combinations} are passed to the Combination Function to enumerate candidate plans, which are then filtered using the \texttt{constraints} by the Filter Function. Finally, the Deliver Function converts the structured solution into a natural language response. The details of SCOPE workflow is presented in Algorithm \ref{alg:pseudocode}.

\begin{table*}[htbp]
\centering
\small
\setlength{\abovecaptionskip}{0.2cm}
\setlength{\belowcaptionskip}{-0.0cm}
\renewcommand{\arraystretch}{0.8}
\begin{adjustbox}{max width=\textwidth}
\begin{tabular}{p{2.5cm}lcccccccccc}
\toprule
\multirow{2}{*}{MODEL} & \multirow{2}{*}{SETTING} 
& \multicolumn{6}{c}{TravelPlanner} 
& \multicolumn{1}{c}{Trip.} 
& \multicolumn{1}{c}{Meet.} \\
\cmidrule(lr){3-8} \cmidrule(lr){9-9} \cmidrule(lr){10-10} 
& & DR & \multicolumn{2}{c}{CPR} & \multicolumn{2}{c}{HCPR} & \textbf{SR} 
& \textbf{SR} & \textbf{SR} \\
\cmidrule(lr){4-5} \cmidrule(lr){6-7}
& & & Micro & Macro & Micro & Macro & 
& & & \\
\midrule

\multirow{4}{*}{GPT-4o}
& Direct       & 100 & 93.6 & 57.0 & 56.4 & 38.7 & 23.6 & 4.9 & 50.7 \\
& CoT \cite{Wei2022ChainOfThought}   & 100 & 95.2 & 67.0 & 64.8 & 45.2 & 31.5 & 3.9 & 47.4 \\
& ToT \cite{Yao2023ToT}      & 95.5 & 87.5 & 48.1 & 64.7 & 47.6 & 26.1 & 1.1 & 34.7 \\
& EvoAgent \cite{Yuan2025Evoagent}       & 100 & 69.7 & 0.0 & 0.0 & 0.0 & 0.0 & 8.6 & 10.8 \\
& HTP \cite{gui2025HTP}      & 100 & 83.5 & 18.9 & 54.3 & 53.8 & 16.1 & 8.3 & 52.9 \\
& ToS \cite{Katz2024Tos}      & 65.2 & 28.7 & 0.0 & 0.0 & 0.0 & 0.0 & 12.5 & 59.8 \\
& CPMPy \cite{kostis2024Cp} & 17.6 & 12.3 & 3.2 & 4.1 & 2.5 & 1.4 & 0.0 & 4.8 \\
& SCOPE       & 100 & \textbf{99.7} & \textbf{97.8} & \textbf{97.6} & \textbf{94.8} & \textbf{93.1} & \textbf{87.1} & \textbf{100} \\
\midrule
\multirow{4}{*}{GPT-o3}
& Direct   & 100 & 96.3 & 71.1 & 65.9 & 66.9 & 66.5 & 78.8 & 70.5 \\
& CoT \cite{Wei2022ChainOfThought}      & 100 & 98.6 & 89.1 & 83.2 & 80.7 & 80.4 & 77.9 & 89.8 \\
& ToT \cite{Yao2023ToT}       & 93.5 & 92.5 & 86.4 & 80.7 & 76.1 & 75.8 & 41.4 & 42.4 \\
& EvoAgent \cite{Yuan2025Evoagent}    & 99.9 & 89.4 & 23.7 & 19.9 & 21.4 & 21.3 & 5.4 & 40.6 \\
& HTP \cite{gui2025HTP}      & 100 & 77.4 & 0.8 & 1.4 & 1.3 & 0.8 & 23.5 & 71.5 \\
& ToS \cite{Katz2024Tos}       & 61.0 & 60.7 & 58.6 & 49.9 & 41.7 & 41.7 & 0.0 & 59.7 \\
& CPMPy \cite{kostis2024Cp} & 5.6 & 3.9 & 1.0 & 1.0 & 0.9 & 0.9 & 11.4 & 17.8 \\
& SCOPE      & 100 & \textbf{99.6} & \textbf{97.5} & \textbf{97.2} & \textbf{94.0} & \textbf{92.2} & \textbf{89.1} & \textbf{99.2}\\
\midrule
\multirow{4}{*}{GPT-5}
& Direct   & 100 & 99.6 & 96.6 & 94.3 & 91.4 & 91.2 & 83.3 & 89.3\\
& CoT \cite{Wei2022ChainOfThought}       & 100 & 99.5 & 96.6 & 94.3 & 91.8 & 91.5 & 84.6 & 93.2 \\
& ToT \cite{Yao2023ToT}     & 99.8 & 99.3 & 95.9 & 90.3 & 85.3 & 85.3 & 61.6 & 55.8 \\
& HTP \cite{gui2025HTP}      & 100 & 82.4 & 14.0 & 18.5 & 17.2 & 11.5 & 5.1 & 67.4 \\
& ToS  \cite{Katz2024Tos}      & 79.5 & 58.1 & 0.0 & 36,7 & 7.4 & 0.0 & 0.0 & 10.6 \\
& CPMPy \cite{kostis2024Cp} & 16.6 & 15.9 & 13.4 & 13.0 & 12.2 & 12.2 & 18.5 & 9.1 \\
& SCOPE    & 100 &  \textbf{99.7} & \textbf{98.3} & \textbf{97.8} & \textbf{95.3} & \textbf{93.9} & \textbf{89.5} & \textbf{100}\\
\midrule
\multirow{4}{*}{Gemini-1.5-Pro} 
& Direct   & 100 & \textbf{96.8} & 75.0 & 54.6 & 27.5 & 21.8 & 38.1 & 43.2 \\
& CoT \cite{Wei2022ChainOfThought}       & 100 & 95.1 & 66.8 & 58.7 & 35.5 & 25.3 & 31.8 & 45.4 \\
& SCOPE       & 97.5 & 95.9 & \textbf{85.6} & \textbf{89.3} & \textbf{79.0} & \textbf{71.6} & \textbf{80.4} & \textbf{90.6} \\
\midrule
\multirow{4}{*}{Gemini-2.5-Pro} 
& Direct   & 100 & 99.5 & 95.7 & 93.8 & 92.5 & 91.7 & 88.8 & 95.2 \\
& CoT \cite{Wei2022ChainOfThought}      & 100 & 99.6 & 96.9 & 96.3 & 94.8 & 93.8 & 88.1 & 96.4 \\
& SCOPE       & 100 & \textbf{99.7} & \textbf{98.2} & \textbf{98.0} & \textbf{95.6} & \textbf{94.2} & \textbf{91.0} & \textbf{99.6} \\
\bottomrule
\end{tabular}
\end{adjustbox}
\caption{Performance comparison across models and settings in inference mode.}
\label{tab:inference_table}
\end{table*}


\begin{figure*}[t]
    \centering
    \setlength{\abovecaptionskip}{0.2cm}
    \setlength{\belowcaptionskip}{-0.3cm}
    \includegraphics[width=1.0\textwidth, height=0.5\textwidth]{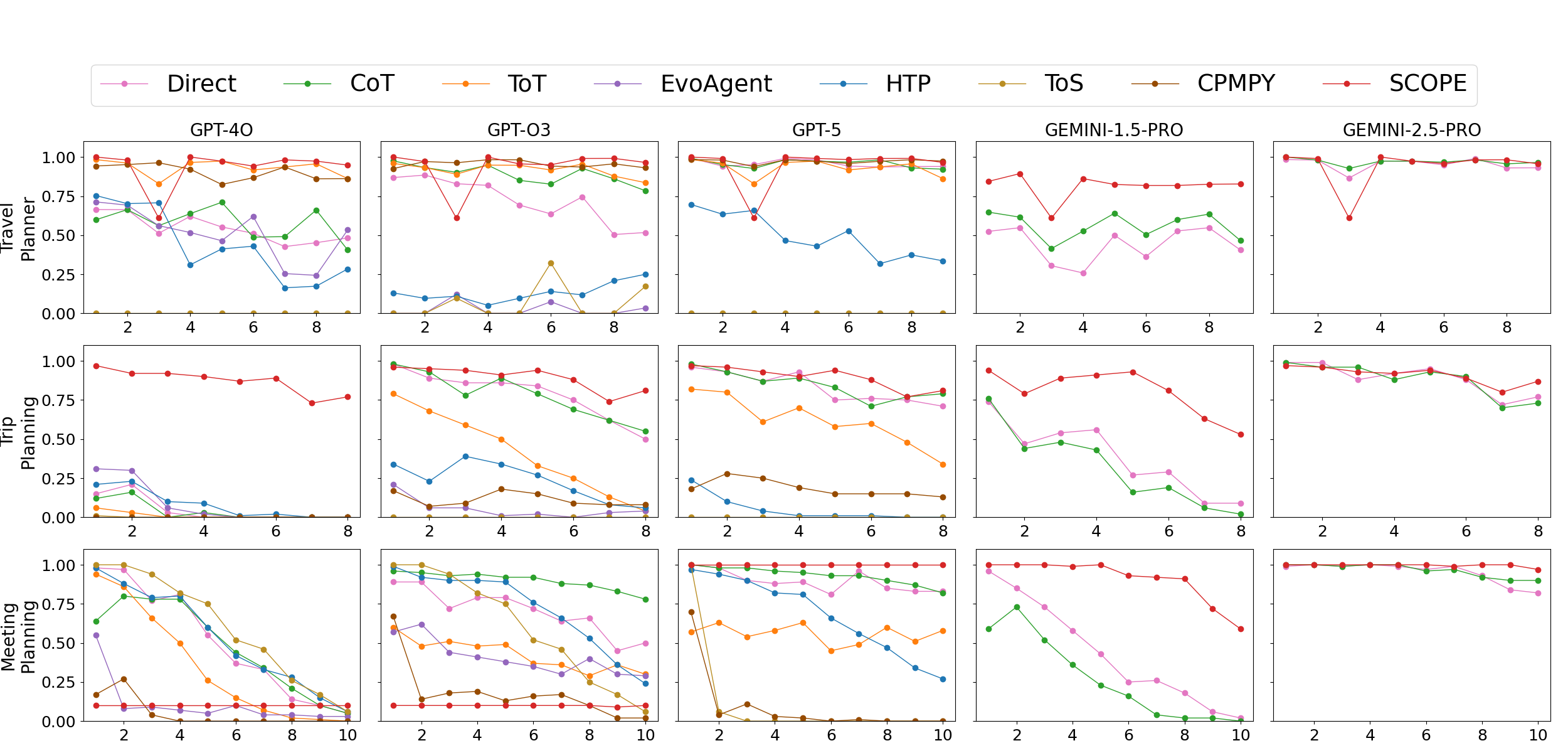} 
    \caption{The success rate (y-axis) across different level of complexity (x-axis). Top to bottom (Different benchmarks): TravelPlanner, Trip Planning, Meeting Planning. Left to right (Different models): GPT-5, GPT-o3, GPT-4o, Gemini-2.5-Pro, Gemini-1.5-Pro.}
    \label{fig:accuracy_over_level}
\end{figure*}

\begin{figure*}[t]
    \centering
    \setlength{\abovecaptionskip}{0.2cm}
    \setlength{\belowcaptionskip}{-0.3cm}
    \includegraphics[width=1.0\textwidth, height=0.5\textwidth]{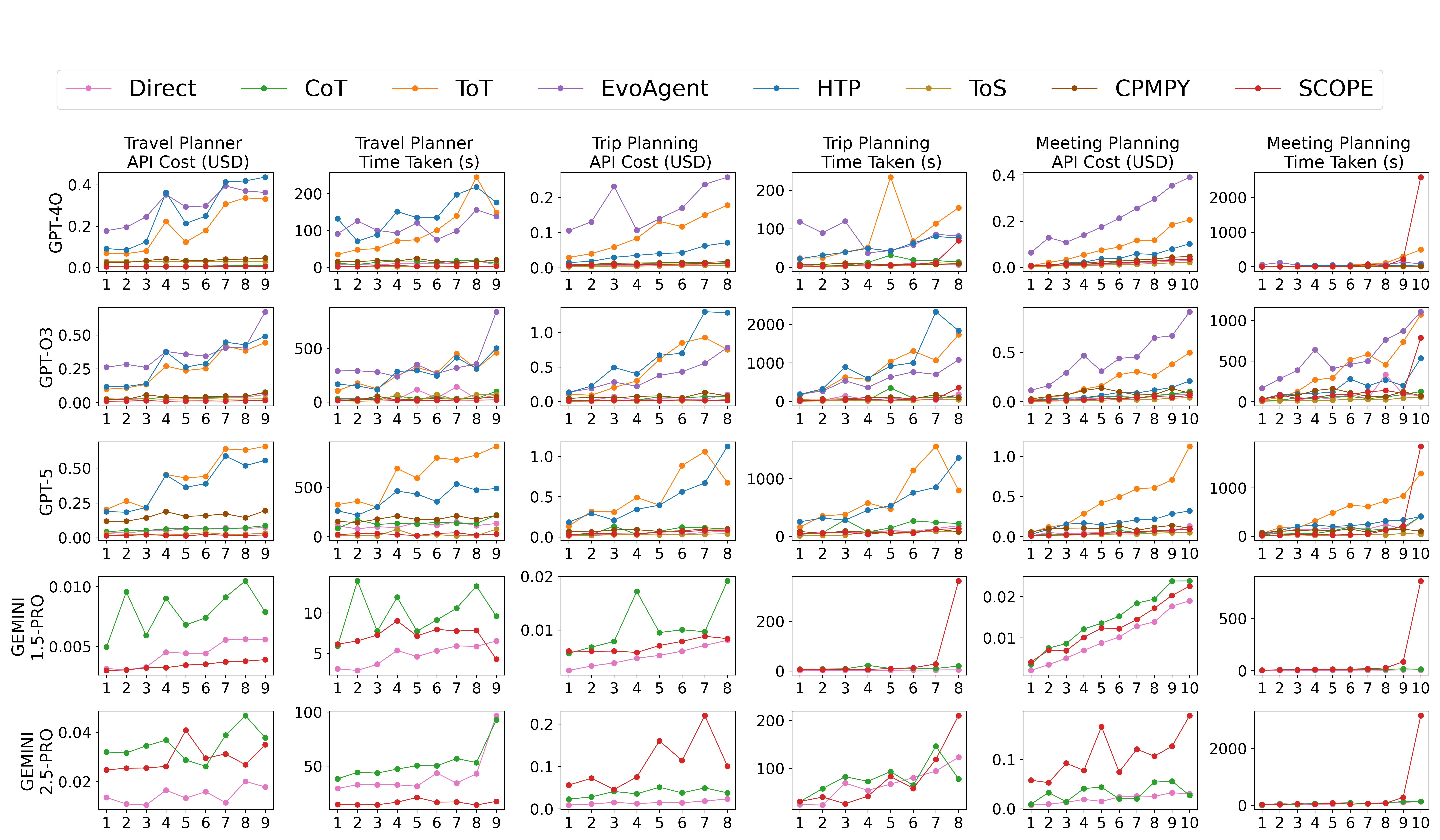}
    \caption{The cost/time required per query (y-axis) for different methods across different level of complexity (x-axis). Top to bottom (Different models): GPT-5, GPT-o3, GPT-4o, Gemini-2.5-Pro, Gemini-1.5-Pro. Left to right (Different metrics): Cost and time taken for TravelPlanner, Trip Planning, and Meeting Planing.}
    \label{fig:all_cost_time}
\end{figure*}

\section{Experiments}

We evaluate SCOPE on \textbf{multi-constraint planning tasks} that require reasoning over combinatorial possibilities and multiple interdependent constraints. We select \textbf{TravelPlanner} \cite{Xie2024TravelPlanner} and \textbf{Natural Plan} \cite{Zheng2024NaturalPlan} for benchmarking, which represent realistic and challenging planning scenarios. We compare our method with 6 baselines across 5 proprietary language models.

\subsection{Datasets}

\textbf{TravelPlanner} is a real-world travel planning benchmark, requiring agent to generate trip itineraries satisfying multiple constraints divided into two types, \textbf{Commonsense Constraints} and \textbf{Hard Constraints}. The details of each type of constraints and metrics are in Appendix \ref{app:travel_planner}. To ensure fairness, all our baselines are provided the same deterministic constraints and operates in closed environment (refer to Appendix \ref{app:fairness}).

\textbf{Natural Plan} is a natural-language planning benchmark where the agent is given query and has to produce the optimal plan. Our experiments focus on two dataset, \textbf{Trip Planning}\footnote{Due to limited budget, we run half of the queries on Trip Planning, with each level of complexity equally sampled. For Direct, CoT and SCOPE, we run full dataset and have a separate table of results at Appendix, Table \ref{tab:inference_table_full_data}. 
} and \textbf{Meeting Planning} of this benchmark. The details of each datasets and metrics are provided in Appendix \ref{app:natural_plan}.

\subsection{Language Models}
We evaluate SCOPE and baselines on two LLMs families: \textbf{GPT} (GPT-5, GPT-o3, and GPT-4o) and \textbf{Gemini} (Gemini-2.5-Pro and Gemini-1.5-Pro). For GPT-5, GPT-o3, and Gemini-2.5-Pro, thinking mode is enabled for enhanced reasoning.  For all models except for GPT-5 and GPT-o3, temperature is set to 0. Temperature for the other two models cannot be set in thinking mode. We use a specific version for each model (see Appendix \ref{app:model_version}).

\subsection{Baselines}
Aside from direct prompting, we include both text-based and solver-based reasoning methods as baseline. The text-based reasoning are Chain-of-Thought (CoT) \cite{Wei2022ChainOfThought}, Tree-of-Thought (ToT) \cite{Yao2023ToT}, EvoAgent \cite{Yuan2025Evoagent} and HyperTree Planning (HTP) \cite{gui2025HTP}. For solver-based reasoning, we use Thought of Search (ToS) \cite{Katz2024Tos} and Constraint Programming and Modeling library in Python (CPMPy) \cite{kostis2024Cp}. The implementation details of each baseline are described in Appendix \ref{app:implementation_details_baselines}. Since SCOPE uses an example query and answer to construct solver functions, all baselines are also given few-shot examples for fair comparison. More details for fairness across baseline are described in Appendix \ref{app:fairness}.

In our experiments, we exclude certain solver-based and code-based baselines which rely on human-provided hints or manual template code \cite{Hao2025Formal, Hao2025Planning}. More justifications can be found in Appendix \ref{app:excluded_solver_baselines}.
The evaluation metrics and the difficulty of different datasets are defined in Appendix \ref{app:travel_planner} for TravelPlanner and Appendix \ref{app:natural_plan} for Natural Plan. However, we further compare against a solver-based approach, \cite{Wang2025PEA}, which is also supported by human-provided hints. SCOPE matches or outperforms this method, demonstrating the effectiveness of automated solver synthesis while maintaining reliable performance. Additional results are provided in Appendix \ref{app:compare_pea}.

\section{Results and Analysis}
\subsection{Overall Success Rate}
Table \ref{tab:inference_table} compares the performances of different methods across different language models\footnote{We choose to selectively implement baselines on Gemini models due to the high cost of baselines and their tendency to fail when following complex instructions.}. The key results are as follows:

\textbf{SCOPE consistently enhances performance across all evaluated models and datasets}. The impact is especially more significant on weaker foundational models like GPT-4o and Gemini-1.5-Pro. For example, the success rate of GPT-4o improves from 12.5\% (ToS) to 87.1\% in Trip Planning, 59.8\% (ToS) to 100\% on Meeting Planning, and 31.5\% (CoT) to 93.1\% on TravelPlanner. These results not only demonstrates SCOPE's strong generalization across diverse models and benchmarks, but also shows SCOPE provides a robust mechanism that enables substantial performance gains on complex tasks where weaker foundational models exhibit limited success.

\textbf{SCOPE enables smaller models to achieve performance comparable to larger models of baselines by leveraging explicit, deterministic logic}. Gains are smaller on very strong models (e.g., GPT-5 and Gemini-2.5-Pro), reflecting their strong reasoning capabilities. On smaller models (GPT-4o, Gemini-1.5-Pro), SCOPE matches or outperforms baselines using larger models at much lower cost (refer to Table \ref{fig:all_cost_time}). For example in Trip Planning, GPT-4o's SCOPE (87.1\%) outperforms the best baseline, CoT of GPT-o3 (78.8\%) and GPT-5 (84.6\%) respectively. This demonstrates that SCOPE's advantages arise from deterministic, executable logic rather than model scale.

\subsection{Success Rate Over Complexity}

We analyze the success rate of baselines and SCOPE across datasets of increasing complexity. The results are visualized in Figure \ref{fig:accuracy_over_level}.

\textbf{SCOPE displays robust performance as problem complexity increases, while baseline methods exhibit substantial performance degradation}. SCOPE consistently maintains a high success rate (often above 90\%) across all complexity levels and datasets, with only a minimal decline at the most extreme levels. In contrast, baseline methods, particularly those relying on weaker foundational models (GPT-4o and Gemini-1.5-Pro), collapses rapidly as difficulty increases. These results shows SCOPE offers superior generalization and scalability compared to existing methods. At inference stage, SCOPE provides parameter values for the solver instead of generating long reasoning chains. This shifts reasoning to parameter estimation for queries, which is less error-prone than reasoning about the full solution. Once the reusable solver function is synthesized, much of the complex reasoning is encoded in deterministic code, further reducing the likelihood of error.

\textbf{SCOPE shows superior usability over long-text reasoning for complex multi-constraint planning tasks}. Baselines that rely on long-text reasoning (e.g., ToT, EvoAgent, and HTP) degrade rapidly as combinatorial complexity increases, due to incomplete enumeration of candidate plans and increased risk to intermediate reasoning errors. Through exhaustive enumeration of candidate plans and logic-based verification, SCOPE maintains stable performance under increasing combinatorial complexity, underscoring its solver-based advantages over text-driven reasoning. This reasoning about parameter values for queries is less error-prone than reasoning about the full solution.

\subsection{Cost and Time Efficiency}

Figure \ref{fig:all_cost_time} presents the inference costs and time of different methods across varying levels of complexity and model types. The summary of corresponding inference costs and latency for each dataset is shown in Appendix, Table \ref{tab:token-cost-time-full}.

\textbf{SCOPE achieves substantial cost efficiency for multi-constraint planning tasks}. Unlike multi-step and multi-agent methods (e.g., ToT, EvoAgent, HTP), which incur exponentially growing API costs due to multiple text generation, SCOPE limits textual output and performs plan enumeration at the code level. This efficient scaling enables it to handle complex, long-horizon planning problems while outperforming iterative reasoning baselines in cost.

\textbf{SCOPE demonstrates a robust trade-off between inference time and solution reliability}. Across all models and datasets, SCOPE requires substantially less inference time than multi-step, long-text reasoning baselines (ToT, EvoAgent, HTP), while remaining comparable to single-agent, single-step (Direct, CoT) and solver-based (ToS) methods. Although single-step methods incur low per-inference cost, SCOPE is more efficient when accounting for the total cost and time needed to obtain a verified correct solution (Figure \ref{fig:cost_req}). While SCOPE’s runtime increases moderately with problem complexity, it consistently produces valid solutions where baselines fail (Figure \ref{fig:accuracy_over_level}), highlighting its scalability and robustness.

\subsection{Error Analysis}
Failures in our method primarily arise from Input Agent. As shown in Appendix \ref{app:capability_error}, smaller models (Gemini-1.5-Pro) may incorrectly assign the value of parameters in the structured representation. Appendix \ref{app:demonstration_error} shows that one-shot demonstration can cause the agent to overgeneralize patterns from examples, resulting in invalid constraint interpretations. Appendix \ref{app:mistakes_error} illustrates that for tasks requiring large structured outputs (Meeting Planning), Input Agent may introduce subtle errors, such as incorrect numerical values, in its output.

\section{Ablation Study}
We conduct an ablation study on the TravelPlanner and Natural Plan tasks by removing individual components of our framework to assess their impact. Specifically, we ablate problem formalization, problem optimization, and solver refinement.

The results from Table \ref{tab:ablation} indicates that each component is essential in the workflow as the absence of component will cause a huge decline in performance. The details of the implementation of ablation is in Appendix \ref{app:ablation_implementation} and analysis of absence of each components is in Appendix \ref{app:ablation_analysis}.

\begin{table}[h]
\centering
\resizebox{\columnwidth}{!}{%
\setlength{\abovecaptionskip}{-0.2cm}
\setlength{\belowcaptionskip}{-0.6cm}
\renewcommand{\arraystretch}{1.0}
\begin{tabular}{lccc}
\toprule
\textbf{Ablation Study} & \textbf{Travel Planner} & \textbf{Trip} & \textbf{Meeting} \\
\midrule
No Problem Formalisation & 25.2 & 13.3 & 100   \\
No Problem Optimisation  & 53.6 & 0.0  & 56.4  \\
No Solver Refinement     & 93.1 & 0.0  & 56.2  \\
SCOPE                    & 93.1 & 87.1 & 100   \\
\bottomrule
\end{tabular}%
}
\caption{Ablation study results from GPT-4o by removing each component from SCOPE workflow.}
\vspace{-0.3cm}
\label{tab:ablation}
\end{table}

\section{Conclusion}
In this work, we introduced SCOPE, a scalable code-based planning framework that disentangles problem reasoning from code execution to address the challenges of multi-constraint complex planning with LLMs. By formalizing natural language queries into structured representations and leveraging reusable solver functions, SCOPE achieves robustness, consistency, and efficiency beyond existing text-based or code-based approaches. Our experiment results show significant improvements in accuracy, cost, and latency, highlighting the effectiveness of systematic, reusable solvers. Our work paves a new way for efficient and robust LLM-based planning in complex real-world tasks.

\section*{Limitations}
While our proposed framework represents a significant step forward in enhancing LLM-based agents' planning abilities, it has some limitations. One limitation is the reliance on LLMs' coding skills and formatted generation capabilities. Hence, our current experiments primarily evaluate closed-source models rather than open-source models. Another limitation is that solver functions are still tied to a specific domain. While they can be reused across different queries within the same domain, they cannot be directly applied to a new domain without redefining the problem structure. Enabling solver generalization across domains is a promising direction for future work. In addition, SCOPE currently does not explicitly handle domain drift, where a query of a new domain cannot be fully expressed within the existing solver’s structural specification. Extending the SCOPE framework with automatic drift detection and adaptive solver regeneration is an important direction.

\section*{Acknowledgement}
This research is supported by the NTU Start-Up Grant, Singapore, and the MOE AcRF Tier 1 Seed Funding Grant (RS37/24), Singapore. Derrick Goh Xin Deik is supported by SINGA scholarship from the Agency for Science, Technology and Research (A*STAR), Singapore.

\bibliography{custom}

\begin{thebibliography}{32}
\providecommand{\natexlab}[1]{#1}

\bibitem[{Achiam et~al.(2023)Achiam, Adler, Agarwal, Ahmad, Akkaya, Aleman, Almeida, Altenschmidt, Altman, Anadkat, and Others}]{achiam2023gpt}
Josh Achiam, Steven Adler, Sandhini Agarwal, Lama Ahmad, Ilge Akkaya, Florencia~Leoni Aleman, Diogo Almeida, Janko Altenschmidt, Sam Altman, Shyamal Anadkat, and Many Others. 2023.
\newblock Gpt-4 technical report.
\newblock \emph{arXiv:2303.08774}.

\bibitem[{Chen et~al.(2024{\natexlab{a}})Chen, Dong, Shu, Zhang, Sesay, Karlsson, Fu, and Shi}]{Chen2024AutoAgents}
Guangyao Chen, Siwei Dong, Yu~Shu, Ge~Zhang, Jaward Sesay, Börje~F. Karlsson, Jie Fu, and Yemin Shi. 2024{\natexlab{a}}.
\newblock Autoagents: A framework for automatic agent generation.
\newblock In \emph{Proceedings of the 33rd International Joint Conference on Artificial Intelligence (IJCAI)}.

\bibitem[{Chen and Li(2024)}]{chen2024toward}
Sijia Chen and Baochun Li. 2024.
\newblock Toward adaptive reasoning in large language models with thought rollback.
\newblock In \emph{Proceedings of the 41st International Conference on Machine Learning (ICML)}.

\bibitem[{Chen et~al.(2024{\natexlab{b}})Chen, Su, Zuo, Yang, Yuan, Chan, Yu, Lu, Hung, Qian, Qin, Cong, Xie, Liu, Sun, and Zhou}]{Chen2024Agentverse}
Weize Chen, Yusheng Su, Jingwei Zuo, Cheng Yang, Chenfei Yuan, Chi-Min Chan, Heyang Yu, Yaxi Lu, Yi-Hsin Hung, Chen Qian, Yujia Qin, Xin Cong, Ruobing Xie, Zhiyuan Liu, Maosong Sun, and Jie Zhou. 2024{\natexlab{b}}.
\newblock Agentverse: Facilitating multi-agent collaboration and exploring emergent behaviors.
\newblock In \emph{Proceedings of the 12th International Conference on Learning Representations (ICLR)}.

\bibitem[{Chen et~al.(2023)Chen, Ma, Wang, and Cohen}]{Chen2023PoT}
Wenhu Chen, Xueguang Ma, Xinyi Wang, and William~W. Cohen. 2023.
\newblock Program of thoughts prompting: Disentangling computation from reasoning for numerical reasoning tasks.
\newblock \emph{Transactions on Machine Learning Research (TMLR)}.

\bibitem[{Dong et~al.(2024)Dong, Li, Dai, Zheng, Ma, Li, Xia, Xu, Wu, Liu, Sun, Li, and Sui}]{dong2024survey}
Qingxiu Dong, Lei Li, Damai Dai, Ce~Zheng, Jingyuan Ma, Rui Li, Heming Xia, Jingjing Xu, Zhiyong Wu, Tianyu Liu, Xu~Sun, Lei Li, and Zhifang Sui. 2024.
\newblock A survey on in-context learning.
\newblock In \emph{Proceedings of the 2024 Conference on Empirical Methods in Natural Language Processing (EMNLP)}.

\bibitem[{Guan et~al.(2023)Guan, Valmeekam, Sreedharan, and Kambhampati}]{guan2023leveraging}
Lin Guan, Karthik Valmeekam, Sarath Sreedharan, and Subbarao Kambhampati. 2023.
\newblock Leveraging pre-trained large language models to construct and utilize world models for model-based task planning.
\newblock In \emph{Proceedings of the 37th International Conference on Neural Information Processing Systems (NeurIPS)}.

\bibitem[{Gui et~al.(2025)Gui, Wang, Wang, Ma, Zhen, Yuan, Hao, Lian, Chen, and Wu}]{gui2025HTP}
Runquan Gui, Zhihai Wang, Jie Wang, Chi Ma, Huiling Zhen, Mingxuan Yuan, Jianye Hao, Defu Lian, Enhong Chen, and Feng Wu. 2025.
\newblock Hypertree planning: Enhancing llm reasoning via hierarchical thinking.
\newblock In \emph{Proceedings of the 42nd International Conference on Machine Learning (ICML)}.

\bibitem[{Han et~al.(2024)Han, Zhang, Yao, Jin, and Xu}]{han2024llm}
Shanshan Han, Qifan Zhang, Yuhang Yao, Weizhao Jin, and Zhaozhuo Xu. 2024.
\newblock Llm multi-agent systems: Challenges and open problems.
\newblock \emph{arXiv:2402.03578}.

\bibitem[{Hao et~al.(2025{\natexlab{a}})Hao, Chen, Zhang, and Fan}]{Hao2025Formal}
Yilun Hao, Yongchao Chen, Yang Zhang, and Chuchu Fan. 2025{\natexlab{a}}.
\newblock Large language models can solve real-world planning rigorously with formal verification tools.
\newblock In \emph{Proceedings of the 2025 Conference of the Nations of the Americas Chapter of the Association for Computational Linguistics (NAACL)}.

\bibitem[{Hao et~al.(2025{\natexlab{b}})Hao, Zhang, and Fan}]{Hao2025Planning}
Yilun Hao, Yang Zhang, and Chuchu Fan. 2025{\natexlab{b}}.
\newblock Planning anything with rigor: General-purpose zero-shot planning with llm-based formalized programming.
\newblock In \emph{Proceedings of the 13th International Conference on Learning Representations (ICLR)}.

\bibitem[{Jiang et~al.(2024)Jiang, Dong, Wang, Fang, Shang, Li, Jin, and Jiao}]{Jiang2024SelfPlanning}
Xue Jiang, Yihong Dong, Lecheng Wang, Zheng Fang, Qiwei Shang, Ge~Li, Zhi Jin, and Wenpin Jiao. 2024.
\newblock Self-planning code generation with large language models.
\newblock \emph{ACM Transactions on Software Engineering and Methodology (TOSEM)}.

\bibitem[{Kambhampati et~al.(2024)Kambhampati, Valmeekam, Guan, Verma, Stechly, Bhambri, Saldyt, and Murthy}]{Kambhampati2024Plan}
Subbarao Kambhampati, Karthik Valmeekam, Lin Guan, Mudit Verma, Kaya Stechly, Siddhant Bhambri, Lucas Saldyt, and Anil Murthy. 2024.
\newblock Position: Llms can’t plan, but can help planning in llm-modulo frameworks.
\newblock In \emph{Proceedings of the 41st International Conference on Machine Learning (ICML)}.

\bibitem[{Katz et~al.(2024)Katz, Kokel, Srinivas, and Sohrabi}]{Katz2024Tos}
Michael Katz, Harsha Kokel, Kavitha Srinivas, and Shirin Sohrabi. 2024.
\newblock Thought of search: Planning with language models through the lens of efficiency.
\newblock In \emph{Proceedings of the 38th Conference on Neural Information Processing Systems (NeurIPS)}.

\bibitem[{Lee et~al.(2025)Lee, Fischer, Wu, Marwood, Baluja, Schuurmans, and Chen}]{lee2025evolving}
Kuang-Huei Lee, Ian Fischer, Yueh-Hua Wu, Dave Marwood, Shumeet Baluja, Dale Schuurmans, and Xinyun Chen. 2025.
\newblock Evolving deeper llm thinking.
\newblock \emph{arXiv:2501.09891}.

\bibitem[{Liu et~al.(2025)Liu, Wang, Sansom, Fu, Choi, Sohn, Kim, and Lee}]{liu2024interactive}
Anthony~Z Liu, Xinhe Wang, Jacob Sansom, Yao Fu, Jongwook Choi, Sungryull Sohn, Jaekyeom Kim, and Honglak Lee. 2025.
\newblock Interactive and expressive code-augmented planning with large language models.
\newblock In \emph{Proceedings of the 63rd Annual Meeting of the Association for Computational Linguistics (ACL)}.

\bibitem[{Madaan et~al.(2023)Madaan, Tandon, Gupta, Hallinan, Gao, Wiegreffe, Alon, Dziri, Prabhumoye, Yang, Gupta, Majumder, Hermann, Welleck, Yazdanbakhsh, and Clark}]{Madaan2023SelfRefine}
Aman Madaan, Niket Tandon, Prakhar Gupta, Skyler Hallinan, Luyu Gao, Sarah Wiegreffe, Uri Alon, Nouha Dziri, Shrimai Prabhumoye, Yiming Yang, Shashank Gupta, Bodhisattwa~Prasad Majumder, Katherine Hermann, Sean Welleck, Amir Yazdanbakhsh, and Peter Clark. 2023.
\newblock Self-refine: Iterative refinement with self-feedback.
\newblock In \emph{Proceedings of the 37th Conference on Neural Information Processing Systems (NeurIPS)}.

\bibitem[{Michailidis et~al.(2024)Michailidis, Tsouros, and Guns}]{kostis2024Cp}
Kostis Michailidis, Dimos Tsouros, and Tias Guns. 2024.
\newblock Constraint modelling with llms using in-context learning.
\newblock In \emph{Proceedings of the 30th International Conference on Principles and Practice of Constraint Programming (CP 2024)}.

\bibitem[{Shinn et~al.(2023)Shinn, Cassano, Gopinath, Narasimhan, and Yao}]{shinn2023reflexion}
Noah Shinn, Federico Cassano, Ashwin Gopinath, Karthik Narasimhan, and Shunyu Yao. 2023.
\newblock Reflexion: Language agents with verbal reinforcement learning.
\newblock In \emph{Proceedings of the 37th International Conference on Neural Information Processing Systems (NeurIPS)}.

\bibitem[{Valmeekam et~al.(2023)Valmeekam, Marquez, Olmo, Sreedharan, and Kambhampati}]{valmeekam2023planbench}
Karthik Valmeekam, Matthew Marquez, Alberto Olmo, Sarath Sreedharan, and Subbarao Kambhampati. 2023.
\newblock Planbench: An extensible benchmark for evaluating large language models on planning and reasoning about change.
\newblock In \emph{Proceedings of the 37th International Conference on Neural Information Processing Systems (NeurIPS)}.

\bibitem[{Wang et~al.(2024)Wang, Mao, Wu, Ge, Wei, and Ji}]{WangSPP2024}
Zhenhailong Wang, Shaoguang Mao, Wenshan Wu, Tao Ge, Furu Wei, and Heng Ji. 2024.
\newblock Unleashing the emergent cognitive synergy in large language models: A task-solving agent through multi-persona self-collaboration.
\newblock In \emph{Proceedings of the 2024 Conference of the North American Chapter of the Association for Computational Linguistics (NAACL)}.

\bibitem[{Wang et~al.(2025)Wang, Weng, Alhanahnah, Jha, and Reps}]{Wang2025PEA}
Zi~Wang, Shiwei Weng, Mohannad Alhanahnah, Somesh Jha, and Tom Reps. 2025.
\newblock Pea: Enhancing llm performance on computational-reasoning tasks.
\newblock \emph{arXiv:2502.10938}.

\bibitem[{Wei et~al.(2025)Wei, Zhang, He, Xia, Pan, and Liu}]{Wei2025PlanGen}
Hui Wei, Zihao Zhang, Shenghua He, Tian Xia, Shijia Pan, and Fei Liu. 2025.
\newblock Plangenllms: A modern survey of llm planning capabilities.
\newblock In \emph{Proceedings of the 63rd Annual Meeting of the Association for Computational Linguistics (ACL)}.

\bibitem[{Wei et~al.(2022)Wei, Wang, Schuurmans, Bosma, Ichter, Xia, H.~Chi, V.~Le, and Zhou}]{Wei2022ChainOfThought}
Jason Wei, Xuezhi Wang, Dale Schuurmans, Maarten Bosma, Brian Ichter, Fei Xia, Ed~H.~Chi, Quoc V.~Le, and Denny Zhou. 2022.
\newblock Chain of thought prompting elicits reasoning in large language models.
\newblock In \emph{Proceedings of the 36th Conference on Neural Information Processing Systems (NeurIPS)}.

\bibitem[{Xie et~al.(2024)Xie, Zhang, Chen, Zhu, Lou, Tian, Xiao, and Su}]{Xie2024TravelPlanner}
Jian Xie, Kai Zhang, Jiangjie Chen, Tinghui Zhu, Renze Lou, Yuandong Tian, Yanghua Xiao, and Yu~Su. 2024.
\newblock Travelplanner: A benchmark for real-world planning with language agents.
\newblock In \emph{Proceedings of the 41st International Conference on Machine Learning (ICML)}.

\bibitem[{Yang et~al.(2025)Yang, Liu, Zhang, Simoulin, Liu, Cao, Teng, Qian, Yang, Luo, and McAuley}]{Yang2025Code2Think}
Dayu Yang, Tianyang Liu, Daoan Zhang, Antoine Simoulin, Xiaoyi Liu, Yuwei Cao, Zhaopu Teng, Xin Qian, Grey Yang, Jiebo Luo, and Julian McAuley. 2025.
\newblock Code to think, think to code: A survey on code-enhanced reasoning and reasoning-driven code intelligence in llms.
\newblock In \emph{Proceedings of the 2025 Conference on Empirical Methods in Natural Language Processing (EMNLP)}.

\bibitem[{Yang et~al.(2023)Yang, Ishay, and Lee}]{Yang2023Coupling}
Zhun Yang, Adam Ishay, and Joohyung Lee. 2023.
\newblock Coupling large language models with logic programming for robust and general reasoning from text.
\newblock In \emph{Findings of the Association for Computational Linguistics (ACL)}.

\bibitem[{Yao et~al.(2023{\natexlab{a}})Yao, Yu, Zhao, Sharfan, L.~Griffiths, Cao, and R~Narasimhan}]{Yao2023ToT}
Shunyu Yao, Dian Yu, Jeffrey Zhao, Izhak Sharfan, Thomas L.~Griffiths, Yuan Cao, and Karthik R~Narasimhan. 2023{\natexlab{a}}.
\newblock Tree of thoughts: Deliberate problem solving with large language models.
\newblock In \emph{Proceedings of the 37th Conference on Neural Information Processing Systems (NeurIPS)}.

\bibitem[{Yao et~al.(2023{\natexlab{b}})Yao, Zhao, Yu, Du, Shafran, Narasimhan, and Cao}]{Yao2023React}
Shunyu Yao, Jeffrey Zhao, Dian Yu, Nan Du, Izhak Shafran, Karthik Narasimhan, and Yuan Cao. 2023{\natexlab{b}}.
\newblock React: Synergizing reasoning and acting in language models.
\newblock In \emph{Proceedings of the International Conference on Learning Representations (ICLR)}.

\bibitem[{Yuan et~al.(2025)Yuan, Song, Chen, Tan, Li, and Yang}]{Yuan2025Evoagent}
Siyu Yuan, Kaitao Song, Jiangjie Chen, Xu~Tan, Dongsheng Li, and Deqing Yang. 2025.
\newblock Evoagent: Towards automatic multi-agent generation via evolutionary algorithms.
\newblock In \emph{Proceedings of the 2025 Conference of the Nations of the Americas Chapter of the Association for Computational Linguistics (NAACL)}.

\bibitem[{Zhang et~al.(2025)Zhang, Goh, Li, Zhang, and Liu}]{cong2024Pmc}
Cong Zhang, Xin~Deik Goh, Dexun Li, Hao Zhang, and Yong Liu. 2025.
\newblock Planning with multi-constraints via collaborative language agents.
\newblock In \emph{Proceedings of the 31st International Conference on Computational Linguistics (COLING)}.

\bibitem[{Zheng et~al.(2024)Zheng, Mishra, Zhang, Chen, Chen, Nova, Hou, Cheng, Le, Chi, and Zhou}]{Zheng2024NaturalPlan}
Huaixiu~Steven Zheng, Swaroop Mishra, Hugh Zhang, Xinyun Chen, Minmin Chen, Azade Nova, Le~Hou, Heng-Tze Cheng, Quoc~V. Le, Ed~H. Chi, and Denny Zhou. 2024.
\newblock Natural plan: Benchmarking llms on natural language planning.
\newblock \emph{arXiv:2406.04520}.

\end{thebibliography}

\clearpage

\appendix
\onecolumn
\section{Pseudocode}
\needspace{6\baselineskip}
\begin{algorithm}[H]
\caption{SCOPE: Structured Solver Construction and Inference}
\label{alg:pseudocode}
\begin{algorithmic}[1]

\State \textbf{Input:} Example query--solution pair $(Q^{ex}, S^{ex})$, test queries $\{Q_i\}_{i=1}^N$
\State \textbf{Output:} Natural-language solutions $\{S_i\}_{i=1}^N$

\vspace{0.5em}
\Comment{\textbf{Stage I: Query-Specific Problem Reasoning}}
\State $(\mathcal{C}, \mathcal{K}) \gets \mathcal{P}(Q^{ex})$ 
\Comment{Planning Agent $\mathcal{P}$: extracts combinations $\mathcal{C}$ and constraints $\mathcal{K}$ from the example query}
\State $\mathcal{S} \gets \boldsymbol{\Sigma}(Q^{ex})$ 
\Comment{Solution Agent $\boldsymbol{\Sigma}$: defines target solution structure $\mathcal{S}$}
\For{each optimization agent $\mathcal{O}_j \in \boldsymbol{\mathcal{O}}$}
    \State $(\mathcal{C}, \mathcal{K}) \gets \mathcal{O}_j(\mathcal{C}, \mathcal{K})$
\EndFor
\Comment{Optimization Agents $\boldsymbol{\mathcal{O}}$: sequentially refine $\mathcal{C}$ and $\mathcal{K}$}

\vspace{0.5em}
\Comment{\textbf{Stage II: Generic Solver Generation}}

\Comment{Combination Solver Generator $\mathcal{G}_{\text{comb}}$: constructs $F_{\text{comb}}$}
\State $F_{\text{comb}} \gets \mathcal{G}_{\text{comb}}(\mathcal{C})$ 
\Comment{Initial construction, no previous output or supervision yet}
\While{$\mathcal{S} \nsubseteq F_{\text{comb}}(\mathcal{C})$}
    \State $F_{\text{comb}} \gets \mathcal{G}_{\text{comb}}\!\left(
        \mathcal{C},
        F_{\text{comb}}(\mathcal{C}),
        \mathcal{S}
    \right)$
    \Comment{Refinement: use previous output + ground-truth structure $\mathcal{S}$}
\EndWhile

\Comment{Filter Solver Generator $\mathcal{G}_{\text{filter}}$: constructs $F_{\text{filter}}$}
\State $F_{\text{filter}} \gets \mathcal{G}_{\text{filter}}(\mathcal{K})$
\Comment{Initial construction, no previous output or supervision yet}
\While{$\mathcal{S} \neq F_{\text{filter}}(F_{\text{comb}}(\mathcal{C}), \mathcal{K})$}
    \State $F_{\text{filter}} \gets \mathcal{G}_{\text{filter}}\!\left(
        \mathcal{K},
        F_{\text{filter}}(F_{\text{comb}}(\mathcal{C}), \mathcal{K}),
        \mathcal{S}
    \right)$
    \Comment{Refinement: use previous output + ground-truth structure $\mathcal{S}$}
\EndWhile

\Comment{Deliver Solver Generator $\mathcal{G}_{\text{deliver}}$: constructs $F_{\text{deliver}}$}
\State $F_{\text{deliver}} \gets \mathcal{G}_{\text{deliver}}(\mathcal{S})$
\Comment{Initial construction, no previous output or supervision yet}
\While{$S^{ex} \neq F_{\text{deliver}}(\mathcal{S})$}
    \State $F_{\text{deliver}} \gets \mathcal{G}_{\text{deliver}}\!\left(
        \mathcal{S},
        F_{\text{deliver}}(\mathcal{S}),
        S^{ex}
    \right)$
    \Comment{Refinement: use previous output + ground-truth solution $S^{ex}$}
\EndWhile

\vspace{0.5em}
\Comment{\textbf{Stage III: Inference}}
\For{$i = 1$ to $N$}
    \State $(\mathcal{C}_i, \mathcal{K}_i) \gets \mathcal{I}(Q_i \mid Q^{ex}, \mathcal{C}, \mathcal{K})$ 
    \Comment{Input Agent $\mathcal{I}$: one-shot demonstration using $Q^{ex}$ and  $(\mathcal{C}, \mathcal{K})$}
    \State $\mathcal{X}_i \gets F_{\text{comb}}(\mathcal{C}_i)$
    \State $\mathcal{Y}_i \gets F_{\text{filter}}(\mathcal{X}_i, \mathcal{K}_i)$
    \State $S_i \gets F_{\text{deliver}}(\mathcal{Y}_i)$
\EndFor

\State \Return $\{S_i\}_{i=1}^N$

\end{algorithmic}
\end{algorithm}
\twocolumn

\section{Experiment Details}
\subsection{Model Versions}
\label{app:model_version}
\begin{enumerate}
    \item GPT-5: gpt-5-2025-08-07
    \item GPT-o3: o3-2025-04-16
    \item GPT-4o: gpt-4o-2024-11-20
    \item Gemini-2.5-Pro: gemini-2.5-pro-preview-05-06
    \item Gemini-1.5-Pro: gemini-1.5-pro-002
\end{enumerate}

\subsection{Fairness Considerations}
\label{app:fairness}
To ensure a fair comparison across methods, several design decisions are enforced. First, few-shot examples are allowed for all methods. For non-solver-based methods such as Direct Prompting, CoT, ToT, and EvoAgent, example queries and their corresponding answers are provided directly in the prompt. For solver-based methods such as ToS and SCOPE, example queries and answers are incorporated into the solver construction pipeline rather than the prompt itself. The examples of query and solution for TravelPlanner are obtained from the train data while the examples of query and solution for Natural Plan tasks are obtained from the few-shot questions.

Second, all code used in solver-based methods is written by the model itself. No human-written code snippets or task-specific hints are provided. Prior work that embeds human-written code or domain-specific hints is not representative of ordinary natural-language inputs, as models are typically given only the query and must generate their own reasoning. Providing human-written code or hints would partially replace model reasoning, making it difficult to attribute performance gains to the model’s own capabilities. Our aim is to evaluate whether the model can independently construct solver functions from natural language queries alone.

Third, all experiments are conducted in a closed environment, meanings agents operate under deterministic constraints. In the Natural Plan task, queries and example answers already define a closed environment, allowing models can deduce explicit internal constraints. 

In contrast, the TravelPlanner task admits many possible interpretations of constraints. For example, rules such as “multiple vehicles are booked if capacity is exceeded” exist in the environment, but the agent cannot infer them from the query alone. Without explicit deterministic constraints, agents may interpret such rules differently, potentially generating inconsistent plans. To ensure tractability and fair evaluation, these constraints are formalized as deterministic constraints and provided across all baselines.

The TravelPlanner environment is formalized using a set of declarative deterministic constraints as follows:

\begin{enumerate}
    \item Travel between locations occurs on odd-numbered days, while staying within the same city occurs on even-numbered days.
    \item For each city stay, exactly one accommodation, one attraction, and three distinct restaurants are required. The same accommodation must be used for two consecutive days within the city.
    \item If an accommodation’s maximum room capacity exceeds the number of travelers, the agent assumes the user will still book the accommodation, potentially reserving multiple rooms of the same type.
    \item If an accommodation requires a minimum number of nights that exceeds the minimum nights specified in the query, the plan is considered invalid.
    \item Flight costs are computed on a per-ticket basis.
    \item A self-driving car is assumed to have a capacity of five people, while a taxi has a capacity of four people. If the number of travelers exceeds vehicle capacity, multiple vehicles of the same type are assumed to be booked.
    \item Combinations of self-driving with flights or taxis are prohibited.
    \item Restaurants must be unique across the entire trip, repeated restaurant visits are not allowed.
\end{enumerate}

\subsection{Implementation Details of Baselines}
\label{app:implementation_details_baselines}
This section provides a high-level overview of the experimental setup used in evaluating SCOPE and baseline methods.

\begin{itemize}
    \item \textbf{Direct Prompting}: The model is presented with the query and few-shot demonstrations, and is required to directly produce the final answer without any explicit reasoning guidance.
    
    \item \textbf{Chain-of-Thought (CoT)} \cite{Wei2022ChainOfThought}: Generates intermediate reasoning steps in a zero-shot manner before producing the final solution, following the standard CoT prompting paradigm.
    
    \item \textbf{Tree-of-Thought (ToT)} \cite{Yao2023ToT}: Explores multiple reasoning trajectories through iterative tree expansion. The branching factor at each step is determined by the model, and heuristic critic is used to select candidate nodes.
    
    \item \textbf{EvoAgent} \cite{Yuan2025Evoagent}: A multi-agent test-time scaling baseline that formulates reasoning as an evolutionary process over a population of agents. Each agent maintains a candidate solution, and iterative mutation and selection are applied across the population based on model-based evaluation, enabling progressive refinement of reasoning trajectories.
    
    \item \textbf{HyperTree Planning (HTP)} \cite{gui2025HTP}: Constructs a hypertree-structured outline to decompose complex tasks into multiple sub-tasks, iteratively refining and expanding the structure to guide the reasoning process.
    
    \item \textbf{Thought of Search (ToS)} \cite{Katz2024Tos}: A code-based reasoning framework that formulates problem solving as an explicit search procedure. The model generates executable code defining a successor function to expand candidate states and a goal test function to evaluate goal states, enabling systematic exploration and pruning of the solution space via programmable search.

    \item \textbf{Constraint Programming and Modeling library in Python (CPMPy)} \cite{kostis2024Cp}: A classical constraint programming approach that formulates the problem as a set of variables and constraints, and solves it using a generic CP solver. It supports high-level modeling in Python and enables integration with different underlying solvers via a unified interface. We follow its official GitHub implementation\footnote{\url{https://github.com/CPMpy/cpmpy}} for evaluation.
    
\end{itemize}

\subsection{Excluded Solver-Based Baselines}
\label{app:excluded_solver_baselines}
We do not include several recent solver-based approaches in our experimental comparison due to fundamental differences in assumptions and evaluation setup:

\begin{itemize}
    \item \textbf{Large Language Models Can Solve Real-World Planning Rigorously with Formal Verification Tools} \cite{Hao2025Formal} : This approach requires human-written example code for the TravelPlanner task, with explicit solver logic provided to the model. Such manual intervention replaces substantial portions of the model’s reasoning process, making it difficult to attribute performance gains to the model itself.
    
    \item \textbf{Planning Anything with Rigor: General-Purpose Zero-Shot Planning with LLM-based Formalized Programming} \cite{Hao2025Planning}: The method, LLMFP assumes task descriptions can be clearly specified in queries, or else it struggles to define problems' goals and constraints. Our tasks are designed to require model to identify and infer internal constraints that are not explicitly stated.
    
\end{itemize}

In contrast, our approach provides no human-written code templates or structured solver hints. All parameters to generate combinations and constraints are inferred by the model from a single query and example. Furthermore, unlike prior approaches that generate instance-specific executable code, our method constructs reusable solver functions shared across all test queries, substantially improving computational efficiency.

\subsection{Exclusion of Additional Text-Based Reasoning and Multi-Agent Baselines}
\label{app:excluded_text_baselines}
We exclude several text-based reasoning baselines and multi-agent baselines. The reasons are as follows :

\begin{itemize}
    \item \textbf{Self-Refine} \cite{Madaan2023SelfRefine}: This method performs iterative refinement using self-generated feedback. The original EvoAgent paper shows that Self-Refine is consistently outperformed by EvoAgent on structured planning and reasoning tasks and shows that Self-Refine only achieves 1.1\% success rate on TravelPlanner with GPT-4.

    \item \textbf{SPP (Self-collaborative Persona Prompting)} \cite{WangSPP2024}: SPP relies on multi-persona self-collaboration to enhance Chain-of-Thought reasoning. However, the original EvoAgent paper shows that it underperforms SPP on planning-intensive tasks and shows that Self-Refine only achieves 0.6\% success rate on TravelPlanner with GPT-4.

    \item \textbf{AgentVerse} \cite{Chen2024Agentverse}, \textbf{AutoAgents} \cite{Chen2024AutoAgents}, and \textbf{Meta-Task Planning (MTP/PMC)} \cite{cong2024Pmc}: These frameworks focus on coordinating multiple agents via a central planning agent (referred as a Recruiter in AgentVerse, a Manager in MTP, and a Planner in AutoAgents), that decomposes tasks and assign subtasks to the generated agents. Conceptually, they are related. Since these methods rely on text-based reasoning, they share the the same limitation as the text-based reasoning baseline in our experiments, which is an excessive combinatorial search space. Hence, the underperformance of those baselines suffices to draw the same conclusion.
    
\end{itemize}

EvoAgent and HyperTree Planning (HTP) are chosen as primary baselines for text-based reasoning as they represent the strongest recent text-based multi-step reasoning methods. The original EvoAgent paper shows EvoAgent surpasses Self-Refine and SPP, while the original HTP paper shows HTP outperforms EvoAgent and MTP. Since our method consistently outperforms HTP and EvoAgent across all tasks, adding other baselines would not change our conclusions.

\section{Details of Ablation Study}
Our evaluation uses GPT-4o as the backbone model. Each version of ablation removes one component from SCOPE, except solver construction. 

\subsection{Implementation of Ablation}
\label{app:ablation_implementation}
\begin{itemize}
    \item \textbf{No Problem Formalization} : The Planning Agent no longer separates parameters into combinations and constraints. Specifically, the Planning Agent is not instructed to output \texttt{combinations} and \texttt{constraints} keys in structured representation but output \texttt{parameters} instead. The Combination and Constraint Function Generator Agent must implicitly decide which parameters are relevant for the function. We set the patience to 3, which is the frequency the model will refine the function and move on if the function is not able to meet the objective, to avoid deadlock.

    \item \textbf{No Problem Optimization} : For Natural Plan, the Optimization Agents that optimize the structured representations are removed from the workflow. The structured representation generated by Planning Agent are given to the Combination and Constraints Function Generator Agent. For TravelPlanner, it removes the modules that automatically filter, prune and package reference data. Due to the resulting explosion of search space, the generated Combination Function produce at most 100{,}000 sampled candidate plans.

    \item \textbf{No Problem Refinement} : The workflow removes the weak supervision process. As long as the functions generated are executable and able to return plans, they are not further optimized and the logic correctness of functions are not assured.
    
\end{itemize}

We set the patience to 3, meaning that the Function Generator Agent attempts to refine its function up to 3 times before proceeding, preventing potential deadlocks when the objective cannot be met.

\subsection{Analysis of Effect of Ablation}
\label{app:ablation_analysis}
The effects of the absence of each component of SCOPE are as follows : 
\begin{itemize}[itemsep=2pt, parsep=0pt, topsep=2pt]
    \item \textbf{Problem Formalization helps Planning Agents to be more precise with the parameters}: By explicitly separating parameters into \texttt{combinations} and \texttt{constraints}, the Planning Agent is less likely to omit information required for candidate generation or constraint filtering. In the Trip Planning task, performance degrades when this separation is absent, as essential constraint parameters may be overlooked. In TravelPlanner, the Combination Function is difficult to be constructed when combination parameters cannot be cleanly isolated from constraints.
    \item \textbf{Problem Optimization helps prevent logical errors in plan construction due to poorly specified parameters}: In the Trip Planning task, the removal of Problem Optimization causes failure to design Combination Function that return any valid plans because the Planning Agent identifies redundant parameters that interfere with correct plan construction. In Meeting Planning, the Planning Agents omits parameters in either \texttt{combinations} or \texttt{constraints}, leading to missing parameters to construct logically correct functions. For TravelPlanner, the absence of an optimization module forces the Combination Function to rely on random plan generation without effective pruning or data restructuring, causing many valid candidate combinations to be missed.
    \item \textbf{Problem Refinement corrects incorrect logic in functions.}: For Trip Planning and Meeting Planning, the initial generated functions are sometimes incorrect. The absence of refinement causing the errors to propagate into inference queries as the same functions are used.
\end{itemize}

\section{Analysis of Solver Construction with Multiple Examples}
\label{app:multi_example_analysis}
SCOPE supports constructing solvers from multiple examples, allowing the union of constraints across examples to be utilized. In our main experiments, the examples generally covers all domain constraints, hence a randomly selected example is sufficient for solver construction.

To study the case where a single example does not provide full coverage, and the impact of using multiple examples in such setting. Experiments are conducted on the Trip Planning using two modified queries for solver construction with GPT-4o. One of the queries has missing flight connectivity constraints, while the other has missing specific-day constraints. Solvers are constructed using each query individually, as well as both queries jointly.

Our results in Table \ref{tab:multi_example} show that using a single incomplete example for solver construction leads to degraded performance, while combining two incomplete examples recovers performance to 87.1\%, matching the performance of solver constructed with one full-constraint example in Table \ref{tab:inference_table}. This demonstrates that SCOPE can leverage multiple examples to improve robustness via union constraint coverage.

\begin{table}[h]
\centering
\resizebox{\columnwidth}{!}{%
\setlength{\abovecaptionskip}{-0.2cm}
\setlength{\belowcaptionskip}{-0.6cm}
\renewcommand{\arraystretch}{1.0}
\begin{tabular}{lc}
\toprule
\textbf{Ablation Study} & \textbf{Success Rate}\\
\midrule
Missing flight connectivity & 28.1 \\
Missing specific-day  & 74.4  \\
Both examples                    & \textbf{87.1} \\
\bottomrule
\end{tabular}%
}
\caption{Success rates of SCOPE given example with incomplete constraint coverage. Combining two examples with union covers all constraints matches the performance of an example covering all constraints.}
\vspace{-0.3cm}
\label{tab:multi_example}
\end{table}

\section{Comparison with PEA}
\label{app:compare_pea}
To further evaluate the effectiveness of SCOPE as an automated solver, we compare SCOPE against Predicate-Enumeration-Aggregation (PEA) \cite{Wang2025PEA}, a solver-based method that relies on human-provided hints. In PEA, constraints, function parameters, and rules for plan construction and validity checking are explicitly defined by human. The results are shown in Table \label{tab:pea}. Despite PEA’s reliance on manually specified constraints and solver structure, SCOPE consistently matches or outperforms it across most settings. This demonstrates that fully automated solver synthesis can achieve strong performance without human-engineered predicates or templates.

\begin{table}[h]
\centering
\resizebox{\columnwidth}{!}{%
\setlength{\abovecaptionskip}{-0.2cm}
\setlength{\belowcaptionskip}{-0.6cm}
\renewcommand{\arraystretch}{1.0}
\begin{tabular}{lccc}
\toprule
\textbf{Method} & \textbf{Travel Planner} & \textbf{Trip} & \textbf{Meeting} \\
\midrule
PEA (GPT-4o)  & 53.1 & 0.0  & 100  \\
SCOPE (GPT-4o)                    & \textbf{93.1} & \textbf{87.1} & 100   \\
PEA (GPT-o3) & 53.4 & 0.0 & 100   \\
SCOPE (GPT-o3)                    & \textbf{92.2} & \textbf{89.1} & 99.2   \\
PEA (GPT-5)  & 53.7 & 0.0  & 100  \\
SCOPE (GPT-5)                   & \textbf{93.1} & \textbf{87.1} & 100   \\
\bottomrule
\end{tabular}%
}
\caption{Comparison of performance of PEA and SCOPE.}
\vspace{-0.3cm}
\label{tab:pea}
\end{table}

\section{Evaluation Details}
Our evaluation uses all 1,000 test data from TravelPlanner, 800 test data (sampled equally across difficulties) from Trip Planning and all 1,000 test data from Meeting Planning.

\subsection{TravelPlanner}
\label{app:travel_planner}
The constraints in TravelPlanner are divided into two types : 

\begin{itemize}[itemsep=2pt, parsep=0pt, topsep=2pt]
    \item \textbf{Commonsense Constraints}: Implicit, commonsense knowledge for generating plausible plans, such as not visiting the same restaurants and attractions throughout the trip.
    \item \textbf{Hard Constraints}: Explicitly mentioned user requirements such as prefers self-driving and wants smoking-free accommodation.
\end{itemize}

Our evaluation uses the same metrics from the original paper as follows :

\begin{itemize}[itemsep=2pt, parsep=0pt, topsep=2pt]
    \item \textbf{Delivery Rate (DR)}: Measures whether an agent successfully output a plan.
    \item \textbf{Commonsense Pass Rate (CPR)}: Evaluates the agent’s ability to generate plans that do not violate commonsense. An example of violation is visiting the same attraction or restaurant in a trip. 
    \item \textbf{Hard Constraint Pass Rate (HCPR)}: Evaluates the agent's ability to meet user's preferences that are explicitly mentioned in each query.
    \item \textbf{Success Rate (SR)}: Evaluates the proportion of generated plans that meet all criteria from commonsense and hard constraint. This is the most important metric as it indicates the agent’s overall effectiveness in producing practical plans.
\end{itemize}

In addition, \textsc{TravelPlanner} computes the scores of CPR and HCPR via two metrics:
\begin{itemize}[itemsep=2pt, parsep=0pt, topsep=2pt]
\item \textbf{Micro}: Computes the fraction of satisfied constraints over the total number of constraints.
\item \textbf{Macro}: Measures the proportion of plans that satisfy all commonsense or all hard constraints.
\end{itemize}

Task difficulty is originally categorized into three levels: \emph{easy}, \emph{medium}, and \emph{hard}. Easy queries primarily involve budget constraints for a single traveler, medium queries introduce additional hard constraints (e.g., cuisine type, room type, or room rules) and vary the number of travelers, hard queries further incorporate transportation preferences and multiple simultaneous constraints.

In our analysis, we further stratify tasks by planning horizon, resulting in nine complexity levels in total. Levels~1--3 correspond to 3-day plans (easy, medium, hard), Levels~4--6 correspond to 5-day plans, and Levels~7--9 correspond to 7-day plans. This setup enables a finer-grained evaluation of performance as both constraint difficulty and planning horizon increase.

\subsection{Natural Plan}
\label{app:natural_plan}
The two datasets of Natural Plan we are evaluating are as follows :  
\vspace{-0.5mm}
\begin{itemize}[itemsep=2pt, parsep=0pt, topsep=2pt]
    \item \textbf{Trip Planning:} 1,600 queries where user requests to design a trip itinerary to travel to multiple cities. The constraints are total number of days in each city, specific days where the user wants to be in a city and available flight connectivity from one city to another.
    \item \textbf{Meeting Planning:} 1,000 queries focused on scheduling meetings that maximize the numbers of friends met. The constraints include the available location and meeting time of friends and distance from one place to another.
\end{itemize}

For both Trip Planning and Meeting Planning tasks, we evaluate model performance using the \textbf{Success Rate} metric like the original paper. Since each query has a single correct plan (gold plan), the success rate is defined as the fraction of generated plans that exactly match the gold plan for the corresponding query.

For Trip Planning, difficulty is defined by the number of cities to visit, increasing from 3 to 8 across 8 levels. For Meeting Planning, difficulty is determined by the number of people to meet, increasing from 1 to 10 across 10 levels. This stratification allows analysis of performance as task complexity increases with the size of the planning problem.

\clearpage

\section{Inference Cost and Efficiency}
\label{app:efficiency}
This appendix presents a comprehensive analysis of the inference cost and efficiency of SCOPE and baseline methods. We report the number of input tokens, the number of output tokens, including intermediate reasoning tokens produced by models operating in thinking modes such as GPT-o3, GPT-5, and Gemini-2.5-Pro, as well as the total inference cost and end-to-end wall-clock inference time. All measurements are conducted under identical experimental settings and evaluation prompts to ensure fair comparison. Token counts include all tokens consumed during inference, including intermediate reasoning tokens where applicable. Inference cost is computed based on the official pricing of each model at the time of evaluation, and inference time is measured as the full wall-clock duration required to generate a final solution, including all agent interactions and tool calls. Results on the full evaluation set indicate that SCOPE maintains superior task performance while achieving competitive inference cost and runtime compared to baseline methods.
\begin{table}[h]
\centering
\small
\begin{adjustbox}{max width=\columnwidth}
\label{tab:model-pricing}
\begin{tabular}{lcc}
\toprule
\textbf{Model} & \textbf{Input Token Price} & \textbf{Output Token Price} \\
\midrule
GPT-5           & 1.25  & 10   \\
GPT-o3          & 2.00  & 8    \\
GPT-4o          & 2.50  & 10   \\
Gemini-2.5-Pro  & 1.25  & 15   \\
Gemini-1.5-Pro  & 1.25  & 10   \\
\bottomrule
\end{tabular}
\end{adjustbox}
\caption{Official model pricing (in USD per 1M tokens).}
\end{table}

\section{Error Analysis}
\subsection{Inference Errors Arising from Mistakes Caused by Limited Instruction-Following Capability}
\label{app:capability_error}
\needspace{6\baselineskip}

\begin{tcblisting}{title=Error from Input Agent in Trip Planning (Gemini-1.5-Pro), listing only, breakable}
Query: You plan to visit 3 European cities for 10 days in total. You only take direct flights to commute between cities. You would like to visit Venice for 6 days. You have to attend a workshop in Venice between day 5 and day 10. You want to spend 2 days in Mykonos. You plan to stay in Vienna for 4 days.

Here are the cities that have direct flights:
Mykonos and Vienna, Vienna and Venice.

Output of Input Agent: {'combinations': {'cities': ['Venice', 'Mykonos', 'Vienna'], 'days_in_city': {'Venice': 6, 'Mykonos': 2, 'Vienna': 2}}, 'constraints': {'direct_flights': [['Mykonos', 'Vienna'], ['Vienna', 'Mykonos'], ['Vienna', 'Venice'], ['Venice', 'Vienna']], 'fixed_days': {'Venice': [[5, 10]]}}}

Error : In 'days_in_city', the value for 'Vienna' should be 4 not 2.
\end{tcblisting}

\subsection{Inference Error Due to Misinterpretation of Demonstration Outputs}
\label{app:demonstration_error}
\needspace{6\baselineskip}

\begin{tcblisting}{title=Error from Input Agent in Trip Planning (GPT-4o), listing only, breakable}
Demonstration Query : You plan to visit 10 European cities for 21 days in total. You only take direct flights to commute between cities. You plan to stay in Vienna for 3 days. Youend 2 days in Oslo. You are going to attend a wedding want to spend 5 days in Frankfurt. You want to sp in Oslo between day 20 and day 21. You want to spend 3 days in Prague. You would like to visit Valencia for 2 days. You want to meet a friend in Valencia between day 17 and day 18. You plan to stay in Dubrovnik for 2 days. You would like to visit Edinburgh for 5 days. From day 1 to day 5, there is a annual show you want to attend in Edinburgh. You plan to stay in London for 2 days. You plan to visit relatives in London between day 12 and day 13. You would like to visit Munich for 3 days. You would like to meet your friends at Munich between day 18 and day 20 to tour together. You would like to visit Budapest for 3 days.

Here are the cities that have direct flights:
Valencia and Munich, Vienna and Munich, Vienna and Valencia, London and Budapest, London and Oslo, Edinburgh and Budapest, Frankfurt and Budapest, Frankfurt and London, Prague and Oslo, Edinburgh and Oslo, Edinburgh and Munich, Prague and Munich, London and Prague, Edinburgh and London, Edinburgh and Frankfurt, Dubrovnik and Munich, Dubrovnik and Vienna, Munich and Oslo, Dubrovnik and Oslo, Budapest and Munich, Frankfurt and Prague, Vienna and London, Frankfurt and Vienna, Frankfurt and Oslo, Frankfurt and Munich, Vienna and Oslo, Vienna and Prague, Budapest and Oslo, Budapest and Prague, London and Valencia, London and Munich, Frankfurt and Dubrovnik, Prague and Valencia, Frankfurt and Valencia, Edinburgh and Prague, Vienna and Budapest.

Demonstration Output : {'combinations': {'cities': ['Edinburgh', 'Frankfurt', 'Dubrovnik', 'Vienna', 'London', 'Budapest', 'Prague', 'Valencia', 'Munich', 'Oslo'], 'direct_flights': [['Valencia', 'Munich'], ['Munich', 'Valencia'], ['Vienna', 'Munich'], ['Munich', 'Vienna'], ['Vienna', 'Valencia'], ['Valencia', 'Vienna'], ['London', 'Budapest'], ['Budapest', 'London'], ['London', 'Oslo'], ['Oslo', 'London'], ['Edinburgh', 'Budapest'], ['Budapest', 'Edinburgh'], ['Frankfurt', 'Budapest'], ['Budapest', 'Frankfurt'], ['Frankfurt', 'London'], ['London', 'Frankfurt'], ['Prague', 'Oslo'], ['Oslo', 'Prague'], ['Edinburgh', 'Oslo'], ['Oslo', 'Edinburgh'], ['Edinburgh', 'Munich'], ['Munich', 'Edinburgh'], ['Prague', 'Munich'], ['Munich', 'Prague'], ['London', 'Prague'], ['Prague', 'London'], ['Edinburgh', 'London'], ['London', 'Edinburgh'], ['Edinburgh', 'Frankfurt'], ['Frankfurt', 'Edinburgh'], ['Dubrovnik', 'Munich'], ['Munich', 'Dubrovnik'], ['Dubrovnik', 'Vienna'], ['Vienna', 'Dubrovnik'], ['Munich', 'Oslo'], ['Oslo', 'Munich'], ['Dubrovnik', 'Oslo'], ['Oslo', 'Dubrovnik'], ['Budapest', 'Munich'], ['Munich', 'Budapest'], ['Frankfurt', 'Prague'], ['Prague', 'Frankfurt'], ['Vienna', 'London'], ['London', 'Vienna'], ['Frankfurt', 'Vienna'], ['Vienna', 'Frankfurt'], ['Frankfurt', 'Oslo'], ['Oslo', 'Frankfurt'], ['Frankfurt', 'Munich'], ['Munich', 'Frankfurt'], ['Vienna', 'Oslo'], ['Oslo', 'Vienna'], ['Vienna', 'Prague'], ['Prague', 'Vienna'], ['Budapest', 'Oslo'], ['Oslo', 'Budapest'], ['Budapest', 'Prague'], ['Prague', 'Budapest'], ['London', 'Valencia'], ['Valencia', 'London'], ['London', 'Munich'], ['Munich', 'London'], ['Frankfurt', 'Dubrovnik'], ['Dubrovnik', 'Frankfurt'], ['Prague', 'Valencia'], ['Valencia', 'Prague'], ['Frankfurt', 'Valencia'], ['Valencia', 'Frankfurt'], ['Edinburgh', 'Prague'], ['Prague', 'Edinburgh'], ['Vienna', 'Budapest'], ['Budapest', 'Vienna']], 'city_stays': {'Edinburgh': 5, 'Frankfurt': 5, 'Dubrovnik': 2, 'Vienna': 3, 'London': 2, 'Budapest': 3, 'Prague': 3, 'Valencia': 2, 'Munich': 3, 'Oslo': 2}}, 'constraints': {'specific_days': {'Edinburgh': [1, 5], 'Oslo': [20, 21], 'Valencia': [17, 18], 'London': [12, 13], 'Munich': [18, 20]}}}

Query: You plan to visit 3 European cities for 15 days in total. You only take direct flights to commute between cities. You want to spend 6 days in Split. You want to spend 4 days in Manchester. You want to spend 7 days in Riga.

Here are the cities that have direct flights:
Riga and Manchester, from Manchester to Split.

Output of Input Agent: {'structured_output': {'combinations': {'cities': ['Split', 'Manchester', 'Riga'], 'direct_flights': [['Riga', 'Manchester'], ['Manchester', 'Riga'], ['Manchester', 'Split'], ['Split', 'Manchester']], 'city_stays': {'Split': 6, 'Manchester': 4, 'Riga': 7}}, 'constraints': {'specific_days': {}}}} 

Error : ['Split', 'Manchester'] should not be one of the 'direct_flight' value
\end{tcblisting}

\subsection{Inference Error Due to Mistakes Induced by Long Outputs}
\label{app:mistakes_error}

\begin{tcblisting}{title=Error from Input Agent in Meeting Planning (GPT-o3), listing only, breakable}
Query: You are visiting San Francisco for the day and want to meet as many friends as possible. Solve the problem by considering various different schedules and picking the best one to optimize your goals.

Travel distances (in minutes): Embarcadero to The Castro: 25. Embarcadero to Chinatown: 7. Embarcadero to Sunset District: 30. Embarcadero to Nob Hill: 10. Embarcadero to Financial District: 5. Embarcadero to Richmond District: 21. The Castro to Embarcadero: 22. The Castro to Chinatown: 20. The Castro to Sunset District: 17. The Castro to Nob Hill: 16. The Castro to Financial District: 20. The Castro to Richmond District: 16. Chinatown to Embarcadero: 5. Chinatown to The Castro: 22. Chinatown to Sunset District: 29. Chinatown to Nob Hill: 8. Chinatown to Financial District: 5. Chinatown to Richmond District: 20. Sunset District to Embarcadero: 31. Sunset District to The Castro: 17. Sunset District to Chinatown: 30. Sunset District to Nob Hill: 27. Sunset District to Financial District: 30. Sunset District to Richmond District: 12. Nob Hill to Embarcadero: 9. Nob Hill to The Castro: 17. Nob Hill to Chinatown: 6. Nob Hill to Sunset District: 25. Nob Hill to Financial District: 9. Nob Hill to Richmond District: 14. Financial District to Embarcadero: 4. Financial District to The Castro: 23. Financial District to Chinatown: 5. Financial District to Sunset District: 31. Financial District to Nob Hill: 8. Financial District to Richmond District: 21. Richmond District to Embarcadero: 19. Richmond District to The Castro: 16. Richmond District to Chinatown: 20. Richmond District to Sunset District: 11. Richmond District to Nob Hill: 17. Richmond District to Financial District: 22.  CONSTRAINTS: You arrive at Embarcadero at 9:00AM. Daniel will be at The Castro from 1:45PM to 5:30PM. You'd like to meet Daniel for a minimum of 120 minutes. Mark will be at Chinatown from 4:45PM to 8:30PM. You'd like to meet Mark for a minimum of 105 minutes. Jeffrey will be at Sunset District from 5:45PM to 8:00PM. You'd like to meet Jeffrey for a minimum of 45 minutes. John will be at Nob Hill from 10:30AM to 3:15PM. You'd like to meet John for a minimum of 90 minutes. David will be at Financial District from 10:30AM to 5:30PM. You'd like to meet David for a minimum of 75 minutes. Robert will be at Richmond District from 2:30PM to 9:00PM. You'd like to meet Robert for a minimum of 105 minutes.

{'combinations': {'origin': {'place': 'Embarcadero', 'time_24h': '09:00'}, 'friends': [{'friend_name': 'Daniel', 'meet_place': 'The Castro', 'available_from': '13:45', 'available_to': '17:30', 'min_minutes': 120}, {'friend_name': 'Mark', 'meet_place': 'Chinatown', 'available_from': '16:45', 'available_to': '20:30', 'min_minutes': 105}, {'friend_name': 'Jeffrey', 'meet_place': 'Sunset District', 'available_from': '17:45', 'available_to': '20:00', 'min_minutes': 45}, {'friend_name': 'John', 'meet_place': 'Nob Hill', 'available_from': '10:30', 'available_to': '15:15', 'min_minutes': 90}, {'friend_name': 'David', 'meet_place': 'Financial District', 'available_from': '10:30', 'available_to': '17:30', 'min_minutes': 75}, {'friend_name': 'Robert', 'meet_place': 'Richmond District', 'available_from': '14:30', 'available_to': '21:00', 'min_minutes': 105}], 'travel_minutes': {'Embarcadero': {'The Castro': 22, 'Chinatown': 5, 'Sunset District': 30, 'Nob Hill': 9, 'Financial District': 4, 'Richmond District': 19}, 'The Castro': {'Embarcadero': 22, 'Chinatown': 20, 'Sunset District': 17, 'Nob Hill': 16, 'Financial District': 20, 'Richmond District': 16}, 'Chinatown': {'Embarcadero': 5, 'The Castro': 20, 'Sunset District': 29, 'Nob Hill': 6, 'Financial District': 5, 'Richmond District': 20}, 'Sunset District': {'Embarcadero': 30, 'The Castro': 17, 'Chinatown': 29, 'Nob Hill': 25, 'Financial District': 30, 'Richmond District': 11}, 'Nob Hill': {'Embarcadero': 9, 'The Castro': 16, 'Chinatown': 6, 'Sunset District': 25, 'Financial District': 8, 'Richmond District': 14}, 'Financial District': {'Embarcadero': 4, 'The Castro': 20, 'Chinatown': 5, 'Sunset District': 30, 'Nob Hill': 8, 'Richmond District': 21}, 'Richmond District': {'Embarcadero': 19, 'The Castro': 16, 'Chinatown': 20, 'Sunset District': 11, 'Nob Hill': 14, 'Financial District': 21}}}, 

'constraints': {'objective': 'maximise_number_of_friends_met', 'origin': {'place': 'Embarcadero', 'time_24h': '09:00'}, 'friends': [{'friend_name': 'Daniel', 'meet_place': 'The Castro', 'available_from': '13:45', 'available_to': '17:30', 'min_minutes': 120}, {'friend_name': 'Mark', 'meet_place': 'Chinatown', 'available_from': '16:45', 'available_to': '20:30', 'min_minutes': 105}, {'friend_name': 'Jeffrey', 'meet_place': 'Sunset District', 'available_from': '17:45', 'available_to': '20:00', 'min_minutes': 45}, {'friend_name': 'John', 'meet_place': 'Nob Hill', 'available_from': '10:30', 'available_to': '15:15', 'min_minutes': 90}, {'friend_name': 'David', 'meet_place': 'Financial District', 'available_from': '10:30', 'available_to': '17:30', 'min_minutes': 75}, {'friend_name': 'Robert', 'meet_place': 'Richmond District', 'available_from': '14:30', 'available_to': '21:00', 'min_minutes': 105}], 'travel_minutes': {'Embarcadero': {'The Castro': 22, 'Chinatown': 5, 'Sunset District': 30, 'Nob Hill': 9, 'Financial District': 4, 'Richmond District': 19}, 'The Castro': {'Embarcadero': 22, 'Chinatown': 20, 'Sunset District': 17, 'Nob Hill': 16, 'Financial District': 20, 'Richmond District': 16}, 'Chinatown': {'Embarcadero': 5, 'The Castro': 20, 'Sunset District': 29, 'Nob Hill': 6, 'Financial District': 5, 'Richmond District': 20}, 'Sunset District': {'Embarcadero': 30, 'The Castro': 17, 'Chinatown': 29, 'Nob Hill': 25, 'Financial District': 30, 'Richmond District': 11}, 'Nob Hill': {'Embarcadero': 9, 'The Castro': 16, 'Chinatown': 6, 'Sunset District': 25, 'Financial District': 8, 'Richmond District': 14}, 'Financial District': {'Embarcadero': 4, 'The Castro': 20, 'Chinatown': 5, 'Sunset District': 30, 'Nob Hill': 8, 'Richmond District': 21}, 'Richmond District': {'Embarcadero': 19, 'The Castro': 16, 'Chinatown': 20, 'Sunset District': 11, 'Nob Hill': 14, 'Financial District': 21}}}}

Error: For the 'travel_minutes' key, 'Embarcadero': {'The Castro': 22, 'Chinatown': 5, 'Sunset District': 30, 'Nob Hill': 9, 'Financial District': 4, 'Richmond District': 19} is incorrect, should be 'Embarcadero': {'The Castro': 25, 'Chinatown': 7, 'Sunset District': 30, 'Nob Hill': 10, 'Financial District': 5, 'Richmond District': 21} instead.

\end{tcblisting}


\section{Prompt for Agents of SCOPE}
\label{app:scope_prompt}
We report the full prompts provided to Planning Agent, Optimization Agents, Solution Agent, Combination Function Generator Agent, Solution Function Generator Agent, and Deliver Function Generator Agent, including role definitions and task-specific instructions. These prompts define the structured representations and function interfaces used throughout the system.
\needspace{6\baselineskip}
\begin{tcblisting}{title=Prompt for Planning Agent, listing only, breakable}
You are a combinations and constraints planning agent. 

You are given few-shot examples queries and structured solution corresponding to the query. Based on the few-shot examples, you have to define them as general constraints planning problems. You identify the ways to generate possible combinations of plans and how to filter those plans to retrieve the solution.
You have three agents to assist you in solving the problem. The three agents are :

1. Input Agent - The agent provides set of elements to generate permutations/combinations and constraints as a structured JSON.
2. Combination Function Generator Agent - The agent creates a generic function called combinations_func() that takes any set of elements provided by Input Agent to generate the list of possible plans.
3. Filter Function Generator Agent - The agent creates a generic function called plan_func() that takes in the constraints provided by Input Agent and any list of possible plans outputted by the function created by Combination Function Generator Agent, to generate the most optimal plan that meets the constraints.

Based on the few-shot examples queries, you should observe what is considered valid possible combinations, and instruct Combination Function Generator Agent to produce those valid possible combinations.
You should identify whether the query is a constraints satisfaction problem (CSP) or optimisation problem. If it is a CSP problem, you should instruct the Filter Function Generator Agent to return the plan if it satisfied all constraints, else if it is an optimisation problem, you should instruct Filter Function Generator Agent to go through all plans that satisfied the constraints return the best one. You can only instruct the agent to either return any valid plan or return the best plan, not both.

Based on the few-shot examples queries, you convert the combination parameters and constraints in few-shot examples into structured output in JSON.
The keys should be "combinations" and "constraints" and respectively.

Then, based on the strucuted output you wrote, your instructions to each agent should cover the following:
1. Input Agent - You instruct the agent on how to identify the set of elements to generate permutations/combinations given a query. You also instruct the agent on how to identify the constraints to tackle such problems. The output format should consist of three keys: "combinations", "constraints" and "solutions", similar to the structured output that you wrote.
2. Combination Function Generator Agent - The "combinations" from the output of Input Agent will be the combinations_func() input. Hence, based on the output format of "combinations", you instruct the agent on creating a function on how to use the keys to generate possible plans.  The function is expected to output list of plans with each plan's format is similar to the few-shot structured solution. This output will be given to plan_func().
3. Filter Function Generator Agent - The "constraints" from the output of Input Agent and output of combinations_func() will be the plan_func() input. Based on the output format of "constraints" and "plan", as well as the type of problem, you instruct the agent on creating a function on how to use the keys from "constraints" to filter the plans from "plan". The function is expected to output a plan that matches the format of few-shot structured solution.

Before providing the structured output, you must output Chain-of-Thought (CoT):
1. First, you must analyse the few-shot examples structured solution. For each key in the solution, you must mention what information are essential in point form. All of the information that you listed must be given to "combinations" in your structured output to generate plans that obey the format of structured solution.
2. Then, you must analyse what are the information that could help filter out all the combination of plans. These information are essential to be given to "constraints" to obtain plan that meet the constraints.

Your output format is as below:

<start_of_COT>
Informations that are essential for generating combinations :
<informations in point form>

Constraints that are essential for filtering plans :
<informations in point form>
<end_of_COT>

<start_of_structured_output>
{
    "combinations": <The set of parameters that generate permutations/combinations>,
    "constraints": <The set of constraints>,
    "combinations_description": <Field description of the value in 'combinations' key. Please describe the keys and values in detail. Please ensure it is written in string and not dict.>,
    "constraints_description": <Field description of 'constraints' key. Please describe the keys and values in detail. Please ensure it is written in string and not dict.>,
}
<end_of_structured_output>

<start_of_planning>
{
    "Input Agent": <Your instructions for Input Agent>,
    "Combination Function Generator Agent": <Your instructions for Combination Function Generator Agent>,
    "Filter Function Generator Agent": <Your instructions for Solutions Function Generator Agent>
}
<end_of_planning>

All of your instructions must be as general as possible. Please ensure your JSON is in correct format and able to be parsed. Do not use "..." to indicate more items.

Few-Shot Examples Query:
<few_shot_examples_query>

Few-Shot Examples Structured Solution:
<few_shot_examples_solution>

Few-Shot Examples Structured Solution Description:
<few_shot_solution_description>

\end{tcblisting}

\begin{tcblisting}{title=Prompt for Optimization Agent (Filter Unnecessary Parameters), listing only, breakable}
You are a Optimization Agent. Your job is to filter out unnecessary parameters of the output of Planning Agent.

You will be given the output of Planning Agent. There are two outputs of Planning Agent:
1. Structured output - The agent will provide the combinations and constraints based on the few-shot examples, as well as their corresponding field description.
2. Planning instructions - The agent will write instructions to "Combination Function Generator Agent" and "Solution Function Generator Agent".

Here is the structured output of Planning Agent:
<structured_output>

Here is the planning instructions of Planning Agent:
<planning>

Your main job is as follows:
1. We define parameter as each key in the "combinations" and "constraints" from structured output. You are required to look at constraints that are not represented in list or JSON.
2. If the parameter has no discriminative power, the parameter has to be removed. We define the parameter with no discriminative power the constraint will either cause all plans to pass or fail. On the other hand, parameter that can cause one plan to fail and other plan to not fail have discriminative power.
3. For example, given the problems produces plan A, B and C, where sum of X for plan A, B and C have the same value, due to plan A,B and C have constant set of items, hence the parameter corresponding to the sum of X should be removed, because it will cause either all plans to pass or fail.
4. If the parameter should be removed, you should remove it in the structure output.
5. Hence, we define the parameter with no discriminative power based on the criteria : The parameter applies uniformly across all of the plans.

You MUST consider each key from BOTH "combinations" and "constraints" of the structured output as your parameters.
Please ensure your JSON is in correct format and able to be parsed. Do not use "..." to indicate more items.

Your output format is as follows:
<start_of_COT>
1. <Parameter Type 1> - <Justify whether the parameter applies/does not apply uniformly across all of the plans>. <Mention if the parameter has/does not have discriminative power>. Hence, it should be <removed/not removed>.
...
N. <Parameter Type N> - <Justify whether the parameter applies/does not apply uniformly across all of the plans. <Mention if the parameter has/does not have discriminative power>. Hence, it should be <removed/not removed>.
<end_of_COT>

<start_of_structured_output>
<Your structured output. The ignored parameters should be removed from the structured output>
<end_of_structured_output>

<start_of_planning>
<your planning>
<end_of_planning>

\end{tcblisting}

\begin{tcblisting}{title=Prompt for Optimization Agent (Fixes Valid Combinations Mistaken for Constraints), listing only, breakable}
You are a Optimization Agent. Your job is to observe the validity of output of the Planning Agent.

You will be given the output of Planning Agent. There are two outputs of Planning Agent:
1. Structured output - The agent will provide the combinations and constraints based on the few-shot examples, as well as their corresponding field description.
2. Planning instructions - The agent will write instructions to "Combination Function Generator Agent" and "Solution Function Generator Agent".

Here is the structured output of Planning Agent:
<structured_output>

Here is the planning instructions of Planning Agent:
<planning>

Your main job is as follows:
1. You are only required to look at the constraints in the "constraints" of structured output, not "combinations".
2. Even if the constraint only applies to each item but not all items, you must also see if similar type of constraint applies to other items. You must properly check if the same type of constraints exists for every item in the combination space.
3. For example, given the combination space has item A, B and C, and the given type of constraint has 3 constraints, where each constraint is applied to A, B and C, then the same type of constraint exists for every item in the combination space.
4. If true, then that constraint is a combination parameter, which should be moved to "combinations" in structured output.

Please ensure your JSON is in correct format and able to be parsed. Do not use "..." to indicate more items.

Your output format is as follows:
<start_of_COT>
1. <Constraint Type 1> - <Justify whether the type of constraint exists/does not exist in every item in the combination space>. Hence, it is <combination parameters/constraint>.
...
N. <Constraint Type N> - <Justify whether the type of constraint exists/does not exist in every item in the combination space>. Hence, it is <combination parameters/constraint>.
<end_of_COT>

<start_of_structured_output>
<Your structured output. The constraints that are combination parameter should be moved to "combinations">
<end_of_structured_output>

<start_of_planning>
<your planning>
<end_of_planning>

\end{tcblisting}

\begin{tcblisting}{title=Prompt for Optimization Agent (Validates and Expands Parameters), listing only, breakable}
You are a Optimization Agent. Your job is to observe the validity of output of the Planning Agent.

You will be given the output of Planning Agent. There are two outputs of Planning Agent:
1. Structured output - The agent will provide the combinations, constraints and structured solution based on the few-shot examples, as well as their corresponding field description.
2. Planning instructions - The agent will write instructions to "Combination Function Generator Agent" and "Solution Function Generator Agent".

Here is the list of type of constraints of Planning Agent:
<constraints>

Here is the structured output of Planning Agent:
<structured_output>

Here is the planning instructions of Planning Agent:
<planning>

Your main job is as follows:
1. We define parameter as each key in the "combinations" and "constraints" from structured output. For each type of parameter, you should check if the type of parameter is expandable. Check if there are any that suggest other parameters should also hold, even if they are not explicitly listed. Rewrite each type of parameters if you think it can be expanded.
2. For example, if pair [A,B] exists, then [B,A] might exist, despite not explicitly stated. Hence, such parameter can be expanded. You output Chain-of-Thought to explain your thought process of checking the claim of each parameter and explain whether the parameter is expandable.
3. You should assume the parameter can be expanded unless there are rules that explicitly stated such expansion cannot be done.

You MUST consider each key from BOTH "combinations" and "constraints" of the structured output as the parameter to analyse.
Please ensure your JSON is in correct format and able to be parsed. Do not use "..." to indicate more items.

Your output format is as follows:
<start_of_COT>
1. <Parameter Type 1> - <Justify whether the type of parameter is expandable>. Hence, it is <expandable/non-expandable>.
...
N. <Parameter Type N> - <Justify whether the type of parameter is expandable>. Hence, it is <expandable/non-expandablet>.
<end_of_COT>

<start_of_structured_output>
<Your structured output. For the type of parameter that it is expandable, you should write the expanded parameter>
<end_of_structured_output>

<start_of_planning>
<Please rewrite the planning instructions in JSON based on the format of the given planning instructions.>
<end_of_planning>

\end{tcblisting}

\begin{tcblisting}{title=Prompt for Solution Agent, listing only, breakable}
You are a Solution Agent. 

You are given a solution. Based on the solution, you have to create a structured output for the solution in list or JSON.
You are also required to provide the description of keys and values of the solution.
Please use as less keys as possible to represent the solution in structured output.

Your output format is as below:

<start_of_structured_output>
{
    "solutions": <The solutions>,
    "solutions_description": <Field description of 'solutions' key. Please describe the keys and values in detail>
}
<end_of_structured_output>

For number-based values, please do not use string. If the value is range based, please use list instead.
Please ensure your JSON is in correct format and able to be parsed. Please do not write the field description for "solutions_description" in dictionary, only string is allowed.


Solution:
<few_shot_examples_solution>

\end{tcblisting}

\begin{tcblisting}{title=Prompt for Combination Function Generator Agent, listing only, breakable}
You are the Combination Function Generator Agent. 

You are given such instructions:
<instructions>

You have to follow the provided instructions to create a function that generate a list of possible combinations of plans. The input parameter of the function 'data', which is a dictionary. You should refer to the provided description of input format, example of input, output format and example of output to design your function.

The description of input format is follows:
<input_description>

An example of input is as follows:
<example_input>

The description of output format for each plan is as follows:
<output_description>

An example of a plan is as follows:
<example_output>

Please use product in Python library itertools to help you in your code. 

The name of the function you will generate is "combinations_func()". Please do not include any example usage or print anything in your code.


WARNING: Only apply instructions that are explicitly stated. Do not assume fairness, balance, or symmetry unless specified. Your function must return a list, not a dict.

Before you write your function code, you must output Chain-of-thought (CoT), which you explain your thought-process and programming logic in order to match the output format provided to you. Do not ever hardcode any information. If such information is not provided, please do not use it.

Your output format is as below:
<start_of_COT>
<your COT>
<end_of_COT>

<start_of_code>
<your combinations_func() code>
<end_of_code>

\end{tcblisting}

\begin{tcblisting}{title=Prompt for Combination Function Reflection Agent, listing only, breakable}
You are a Reflection Agent. Your job is to observe the mistake of the output of the Combination Function Generator Agent.

The job of Combination Function Generator Agent is to create a function called combinations_func() that generate a list of possible combinations of plans. The input parameter of the function 'data', which is a dictionary.
However, the agent made a mistake in the function that causes the function to not able to generate combinations of plans that contain the ground-truth plan.

Here is the one of the output plan from the list of plans provided by the function:
<example_input>

Here is the expected ground-truth plan that the generated combinations of plans should contain:
<example_output>

Here is the Chain-of-Thought (CoT) thought process of the agent before writing the code:
<chain_of_thought>

Here is the code written by the agent:
<code>

Your main job is as follows: 
1. You observe the pattern of ground-truth plan and the pattern of one of the output plan from the list of plans. You mention the difference in pattern in your CoT.
2. You should also mention in your CoT that how you should change the CoT thought process of agent to ensure the function generates plan with consistent pattern as the ground-truth plan. 
3. You correct the CoT of the agent and write it in your output. Then, you made the corresponding correction on the agent's code.
4. Please do not hardcode the function to append the ground-truth plan to the solution. Please do not check the combinations against the ground-truth plan. This defeat the purpose of ensuring the function generates valid plans. Similarly like the original code, the final function should output a list, not a dict. Do not use wrapper like ```python``` when writing code. Use the wrapper in the provided output format.

Your output format is as follows:
<start_of_COT>
<your COT>
<end_of_COT>

<start_of_COT_correction>
<corrected CoT of Combination Function Generator Agent>
<end_of_COT_correction>

<start_of_code_correction>
<corrected code of Combination Function Generator Agent>
<end_of_code_correction>


\end{tcblisting}

\begin{tcblisting}{title=Prompt for Filter Function Generator Agent, listing only, breakable}
You are the Filter Function Generator Agent. 

You are given such instructions:
<instructions>

You have to follow the provided instructions to create a function that retrieve the desired plan. The input parameter of the function are 'data' and 'constraints'. 'Data' contains a list of plan, with each plan in JSON format. 'Constraints' contains the constraints. You should refer to the provided description of input plan format, example of input, description of constraints format and example of constraints to design your function. Your function will return one of the plans as solution.

You are given a list of plan, the description of input format for each plan is as follows:
<input_description>

An example of a plan is as follows:
<example_input>

You are also given the constraints, the description of constraints for is as follows:
<constraints_description>

An example of constraints is as follows:
<example_constraints>

The name of the function you will generate is "plan_func()". Please do not include any example usage or print anything in your code.

Remember, the input parameter "data" is a list, while the input parameter "constraints" is JSON.

Your output is a Chain-of-thought (CoT) and Python function as below:
<start_of_COT>
<your COT>
<end_of_COT>

<start_of_code>
<your plan_func() code>
<end_of_code>
\end{tcblisting}

\begin{tcblisting}{title=Prompt for Filter Function Reflection Agent, listing only, breakable}
You are a Reflection Agent. Your job is to observe the mistake of the output of the Filter Function Generator Agent.

The job of Filter Function Generator Agent is to create a function called plan_func() that returns the valid plan given a list of possible combinations of plans. The input parameter of the function are 'data' and 'constraints'. 'Data' contains a list of plan, with each plan in JSON format. 'Constraints' contains the constraints. You should refer to the provided description of input plan format, example of input, description of constraints format and example of constraints to design your function.
However, the agent made a mistake in the function that causes the function to not able to return the ground-truth plan from the possible combinations of plans.

Here is the instructions given to the agent:
<instructions>

Here is the expected ground-truth plan that the function should return:
<example_output>

Here is the plan that the function returns:
<function_output>

Here is the Chain-of-Thought (CoT) thought process of the agent before writing the code:
<chain_of_thought>

Here is the code written by the agent:
<code>

Your main job is as follows: 
1. You observe the difference between ground-truth plan and the plan that the function returns.
2. If the function returns None, it indicates a constraint is handled incorrectly in the code that some plans might meet it but overlooked by the function. Check whether is there possibility that each constraint can be handled differently that some plans can meet such constraints. 
3. If the function does not return None but returns a different plan than ground-truth, you should reflect whether some constraints are written too lenient that some plans might not meet the constraints but overlooked by the function.
4. Once you identified the constraint that is handled incorrectly, you should mention in your CoT that how you should change the CoT thought process of agent to ensure the function handles the constraints correctly. 
5. You correct the CoT of the agent and write it in your output. Then, you made the corresponding correction on the agent's code. 
6. Please do not hardcode the function to return the ground-truth plan to the solution. This defeat the purpose of ensuring the function returns the correct plan. Similarly like the original code, the final function should output a JSON. Do not use wrapper like ```python``` when writing code. Use the wrapper in the provided output format.


Your output format is as follows:
<start_of_COT>
<your COT>
<end_of_COT>

<start_of_COT_correction>
<corrected CoT of Filter Function Generator Agent>
<end_of_COT_correction>

<start_of_code_correction>
<corrected code of Filter Function Generator Agent>
<end_of_code_correction>

\end{tcblisting}

\begin{tcblisting}{title=Prompt for Deliver Function Generator Agent, listing only, breakable}
You are the Deliver Function Generator Agent. 

Given an input, you have to create a function that processes the input and provides output that matches the format of few-shot examples solution. The input parameter of the function 'data', which is a plan in JSON. You should refer to the provided description of input plan format and example of input to design your function.

You are given a plan, the description of a plan is as follows:
<input_description>

An example of a plan is as follows:
<example_input>

The few-shot example solution is as follows:
<few_shot_examples_solution>

The name of the function you will generate is "deliver_func()". Please do not include any example usage or print anything in your code.

Before writing your code, you must output Chain-of-thought (CoT):
1. You must first analyse the format of few-shot example solution. Does it contain any heading? What does the pattern of output look like?
2. Then, you explain how are you going to design the function to ensure the format of your function's output is consistent with the few-shot example solution. Do not miss any heading if the example solution contains heading.

Your output format is CoT and Python function as below:
<start_of_COT>
<your COT>
<end_of_COT>

<start_of_code>
<your deliver_func() code>
<end_of_code>
\end{tcblisting}

\begin{tcblisting}{title=Prompt for Deliver Function Reflection Agent, listing only, breakable}
You are a Reflection Agent. Your job is to observe the mistake of the output of the Deliver Function Generator Agent 

The job of Deliver Function Generator Agent us to create a function that processes a structured input in JSON and provides output that matches the format of few-shot examples solution in natural language. The input parameter of the function 'data', which is a plan in JSON.
However, the agent made a mistake in designing the function, resulting in the output of function does not match the format of few-shot examples solution in natural language, despite the structured input is correct.
The name of the function you will generate is "deliver_func()". The input parameter of function is "data", which is structured input.

Here is the structured input given to the function:
<structured_input>

Here is the expected ground-truth plan from the few-shot examples:
<example_output>

Here is the plan that the function returns:
<current_output>

Here is the code written by the agent:
<code>

Your main job is as follows: 
1. You observe the pattern of ground-truth plan and the pattern of plan that the function returns. You mention the difference in pattern in your CoT.
2. You should also mention in your CoT that how you should change the code to ensure the function generates plan with consistent pattern as the ground-truth plan. 
3. You correct the CoT of the agent and write it in your output. Then, you made the corresponding correction on the agent's code.
4. Please do not hardcode the function to return the ground-truth plan from few-shot examples. Please do not check the output of function against ground-truth plan from few-shot examples. This defeat the purpose of ensuring the function generates plans with valid format. Similarly like the original code, the final function should be a string. Do not use wrapper like ```python``` when writing code. Use the wrapper in the provided output format.

You must ensure the format of your function's output is consistent with the example solution. Do not miss any header if the example solution contains header.

Your output is a Chain-of-thought (CoT) and Python function as below:
<start_of_COT>
<your COT>
<end_of_COT>

<start_of_code>
<your deliver_func() code>
<end_of_code>
\end{tcblisting}

\begin{tcblisting}{title=Prompt for Input Agent, listing only, breakable}
You are the Input Agent. 

You are given such instructions:
<instructions>
The description of output format is as follows:
<output_description>

An example of query is given below:
<example_input>

An example of output is given below:
<example_output>

You will be given a query. You have to generate a strcutured output in JSON adhering to the provided instructions and output format. You should refer to the provided description of output format and example of output. 


Your output format is as below:
<start_of_structured_output>
<your structured output>
<end_of_structured_output>

Query :
<query>

\end{tcblisting}

\clearpage
\section{Prompt for Baselines}
\label{sec:baseline_prompt}
We report the prompts for Tree-of-Thought (ToT) and Thought of Search (ToS), as these methods rely on prompt-level control of reasoning and search. For EvoAgent, we use the prompt provided in the official GitHub repository, with minimal modifications to adapt to different task domains. For HyperTree Planning, we directly adopt the prompt described in the original paper.
\subsection{Prompts for Tree-of-Thought (ToT)}
\needspace{6\baselineskip}

\begin{tcblisting}{title=Prompt for Expansion, listing only, breakable}
Here is an example of query and full solution.

Example query : 
You plan to visit 10 European cities for 21 days in total. You only take direct flights to commute between cities. You plan to stay in Vienna for 3 days. You want to spend 5 days in Frankfurt. You want to spend 2 days in Oslo. You are going to attend a wedding in Oslo between day 20 and day 21. You want to spend 3 days in Prague. You would like to visit Valencia for 2 days. You want to meet a friend in Valencia between day 17 and day 18. You plan to stay in Dubrovnik for 2 days. You would like to visit Edinburgh for 5 days. From day 1 to day 5, there is a annual show you want to attend in Edinburgh. You plan to stay in London for 2 days. You plan to visit relatives in London between day 12 and day 13. You would like to visit Munich for 3 days. You would like to meet your friends at Munich between day 18 and day 20 to tour together. You would like to visit Budapest for 3 days.

Here are the cities that have direct flights:
Valencia and Munich, Vienna and Munich, Vienna and Valencia, London and Budapest, London and Oslo, Edinburgh and Budapest, Frankfurt and Budapest, Frankfurt and London, Prague and Oslo, Edinburgh and Oslo, Edinburgh and Munich, Prague and Munich, London and Prague, Edinburgh and London, Edinburgh and Frankfurt, Dubrovnik and Munich, Dubrovnik and Vienna, Munich and Oslo, Dubrovnik and Oslo, Budapest and Munich, Frankfurt and Prague, Vienna and London, Frankfurt and Vienna, Frankfurt and Oslo, Frankfurt and Munich, Vienna and Oslo, Vienna and Prague, Budapest and Oslo, Budapest and Prague, London and Valencia, London and Munich, Frankfurt and Dubrovnik, Prague and Valencia, Frankfurt and Valencia, Edinburgh and Prague, Vienna and Budapest.
Find a trip plan of visiting the cities for 21 days by taking direct flights to commute between them.

Example full solution ：
Here is the trip plan for visiting the 10 European cities for 21 days:

**Day 1-5:** Arriving in Edinburgh and visit Edinburgh for 5 days.
**Day 5:** Fly from Edinburgh to Frankfurt.
**Day 5-9:** Visit Frankfurt for 5 days.
**Day 9:** Fly from Frankfurt to Dubrovnik.
**Day 9-10:** Visit Dubrovnik for 2 days.
**Day 10:** Fly from Dubrovnik to Vienna.
**Day 10-12:** Visit Vienna for 3 days.
**Day 12:** Fly from Vienna to London.
**Day 12-13:** Visit London for 2 days.
**Day 13:** Fly from London to Budapest.
**Day 13-15:** Visit Budapest for 3 days.
**Day 15:** Fly from Budapest to Prague.
**Day 15-17:** Visit Prague for 3 days.
**Day 17:** Fly from Prague to Valencia.
**Day 17-18:** Visit Valencia for 2 days.
**Day 18:** Fly from Valencia to Munich.
**Day 18-20:** Visit Munich for 3 days.
**Day 20:** Fly from Munich to Oslo.
**Day 20-21:** Visit Oslo for 2 days.


You are a Tree-of-Thought (ToT) Agent. Given a query, a current state and the reasoning up to that state, you have to provide the next step for the current state.
In each step, you will output you next city you would like to visit and provide reasoning.

You must provide the reasoning of choosing this city. In every state, you choose N number of most possible cities (value of N is your choice), that you would like to visit from that current state.
If you are at first step, format is as below:
**Day 1-N:** Arriving in <first city> and visit <first city> for N days.

If you are at second step onwards, format is as below:
**Day X:** Fly from <current city> to <next city>.
**Day X-Y:** Visit <next city> for <N> days.

For example, the current state is:
Here is the trip plan for visiting the 10 European cities for 21 days:

**Day 1-5:** Arriving in Edinburgh and visit Edinburgh for 5 days.

Then, a possible example output of your next step is visiting Frankfurt:
**Day 5:** Fly from Edinburgh to Frankfurt.
**Day 5-9:** Visit Frankfurt for 5 days.



Your output format is a list as follows :
<start_of_output>
[
    {"next_step": <the output for your next step (choosing city 1)>, "reasoning": <your reasoning for choosing city 1>},
    {"next_step": <the output for your next step (choosing city 2)>, "reasoning": <your reasoning for choosing city 2>},
    ...
    {"next_step": <the output for your next step (choosing city N)>, "reasoning": <your reasoning for choosing city N>}
]
<end_of_output>

Query:
<query>

Current state:
<current_state>

Reasoning up to the state:
<reasoning_state>

\end{tcblisting}

\begin{tcblisting}{title=Prompt for Evaluation, listing only, breakable}
You are a Tree-of-Thought (ToT) Evaluator. Given a query, list of current states and reasoning for each state, you have to evaluate and provide score for each state of how likely each state will lead to correct solution.
Your score should be between 1 and 10. Only integer is allowed.
In your output, you must copy each state and reasoning exactly as the input, do not change any words.
You must ensure the score for each output are different. You are not allowed to provide same score for different output.

The current states and reasoning is given in such format:
[
    {"current_state": <current state>, "reasoning": <reasoning for the state>},
    {"current_state": <current state>, "reasoning": <reasoning for the state>},
    {"current_state": <current state>, "reasoning": <reasoning for the state>}
]

Your output format is a list as follows :
<start_of_output>
[
    {"current_state": <current state>, "reasoning": <reasoning for the state>, "score": <integer between 1 and 10>},
    {"current_state": <current state>, "reasoning": <reasoning for the state>, "score": <integer between 1 and 10>},
    {"current_state": <current state>, "reasoning": <reasoning for the state>, "score": <integer between 1 and 10>}
]
<end_of_output>

Query:
<query>

Current states and reasoning:
<current_state>
\end{tcblisting}

\newpage

\subsection{Prompts for Thought of Search (ToS)}
\needspace{6\baselineskip}

\begin{tcblisting}{title=Prompt for Introduction, listing only, breakable}
You are a task description agent. There is a task called <task>. You are given an example query and answer of the task. Using this query and answer, you must provide a detailed description of task.
You must also provide any assumptions or constraints that are not explicit in the query. Your assumptions and constraints must be general so that can be applied on other queries of <task>.
Finally, you must also mention how a state is represented. You should also describe how to transition from one state to another and what is the ending state.
Lastly, you list out the conditions where a state has met the goal by writing the list of conditions after.

Your output format is as below:
Description of <task>:
<your description of task>

Description of how to transition from one state to another and ending state:
<the description>

List of conditions where a state has met the goal:
1. <Condition 1>
...
N. <Condition N>

Here is an example task called 24 Game and here is the example output description :

The 24 Game is a mathematical card game where the objective is to manipulate four integers so that the final result equals 24. The game begins with a list of four numbers, and the player must use each number exactly once, applying any combination of addition, subtraction, multiplication, or division to obtain 24.
A state is represented as a list of numbers (of length four or fewer).


An example query of the task :
<task_query>

An example answer of the task :
<task_answer>

Your output :
\end{tcblisting}

\begin{tcblisting}{title=Prompt for Evaluation, listing only, breakable}
You are a function generator agent. There is a task called <task>.

First of all, you must define the state. To do that, you provide the state from the given example query.

Write a Python function called 'successor_states(state)' that generates all valid successor states for the <task>.

Then, write a Python function called 'is_goal(state)' that tests whether a state is a goal state. 

Then, write a Python function called 'convert_state_to_nl(state)' that convert the goal state to NL that shares the same format as the example answer.
Hence, you must ensure the goal state contain the sufficient information be converted to NL.

Please do not include any example usage or print anything in your code. Ensure your code is written within the given wrapper. Please don't include ```python and ``` wrapper in your output.
Please do not hardcode any fixed values in any functions that cannot be applied to other queries of <task>.


The task description of <task> is as follows :
<task_description>

An example query of the task :
<task_query>

An example answer of the task :
<task_answer>

Your output format is as follows :

State structure:
<start_of_JSON>
<your state given the example query>
<end_of_JSON>

Output code for successor_states:
<start_of_code>
<your code. Please do not hardcode any fixed values in any functions that cannot be applied to other queries.>
<end_of_code>

Output code for is_goal:
<start_of_code>
<your code. Please do not hardcode any fixed values in any functions that cannot be applied to other queries.>
<end_of_code>

Output code for convert_state_to_nl:
<start_of_code>
<your code. Remember, the function output must share the same format as the answer of the task.>
<end_of_code>

\end{tcblisting}

\section{Output of Agents of SCOPE}
\label{sec:output}
We present the outputs of each agent in the SCOPE workflow, using Trip Planning as a case study.
\subsection{Output of Agents}

\begin{tcblisting}{title=Output of Planner Agent, listing only, breakable}
<start_of_COT>
Informations that are essential for generating combinations :
- List of cities to visit.
- Total number of days for the trip.
- Direct flight connections between cities.

Constraints that are essential for filtering plans :
- Specific number of days to stay in each city.
- Specific days to be in certain cities for events or meetings.
- Total trip duration must be exactly 21 days.
- Must only use direct flights between cities.
<end_of_COT>

<start_of_structured_output>
{
    "combinations": {
        "cities": ["Edinburgh", "Frankfurt", "Dubrovnik", "Vienna", "London", "Budapest", "Prague", "Valencia", "Munich", "Oslo"],
        "total_days": 21,
        "direct_flights": [
            ["Valencia", "Munich"], ["Vienna", "Munich"], ["Vienna", "Valencia"], ["London", "Budapest"],
            ["London", "Oslo"], ["Edinburgh", "Budapest"], ["Frankfurt", "Budapest"], ["Frankfurt", "London"],
            ["Prague", "Oslo"], ["Edinburgh", "Oslo"], ["Edinburgh", "Munich"], ["Prague", "Munich"],
            ["London", "Prague"], ["Edinburgh", "London"], ["Edinburgh", "Frankfurt"], ["Dubrovnik", "Munich"],
            ["Dubrovnik", "Vienna"], ["Munich", "Oslo"], ["Dubrovnik", "Oslo"], ["Budapest", "Munich"],
            ["Frankfurt", "Prague"], ["Vienna", "London"], ["Frankfurt", "Vienna"], ["Frankfurt", "Oslo"],
            ["Frankfurt", "Munich"], ["Vienna", "Oslo"], ["Vienna", "Prague"], ["Budapest", "Oslo"],
            ["Budapest", "Prague"], ["London", "Valencia"], ["London", "Munich"], ["Frankfurt", "Dubrovnik"],
            ["Prague", "Valencia"], ["Frankfurt", "Valencia"], ["Edinburgh", "Prague"], ["Vienna", "Budapest"]
        ]
    },
    "constraints": {
        "city_stays": {
            "Edinburgh": 5,
            "Frankfurt": 5,
            "Dubrovnik": 2,
            "Vienna": 3,
            "London": 2,
            "Budapest": 3,
            "Prague": 3,
            "Valencia": 2,
            "Munich": 3,
            "Oslo": 2
        },
        "specific_days": {
            "Edinburgh": [1, 5],
            "Oslo": [20, 21],
            "Valencia": [17, 18],
            "London": [12, 13],
            "Munich": [18, 20]
        }
    },
    "combinations_description": "The 'combinations' key contains the parameters for generating possible travel plans. 'cities' is a list of cities to visit, 'total_days' is the total number of days for the trip, and 'direct_flights' is a list of pairs of cities that have direct flights between them.",
    "constraints_description": "The 'constraints' key contains the conditions that must be met for a valid travel plan. 'city_stays' specifies the number of days to stay in each city, and 'specific_days' specifies the exact days certain cities must be visited for events or meetings."
}
<end_of_structured_output>

<start_of_planning>
{
    "Input Agent": "Identify the set of cities to visit, the total number of days for the trip, and the direct flight connections between cities. Also, identify the constraints such as the number of days to stay in each city and specific days to be in certain cities for events or meetings.",
    "Combination Function Generator Agent": "Create a function combinations_func() that takes 'cities', 'total_days', and 'direct_flights' as input. Generate all possible sequences of city visits that fit within the total days and only use direct flights.",
    "Solution Function Generator Agent": "Create a function plan_func() that takes 'city_stays' and 'specific_days' as constraints, along with the list of possible plans from combinations_func(). Filter the plans to find one that satisfies all constraints, ensuring the total trip duration is exactly 21 days and all specific day requirements are met."
}
<end_of_planning>
\end{tcblisting}

\begin{tcblisting}{title=Output of Optimization Agent (Filter Unnecessary Parameters), listing only, breakable}
<start_of_COT>
1. cities - The parameter does not apply uniformly across all of the plans as different plans can include different subsets of cities. The parameter has discriminative power. Hence, it should not be removed.

2. total_days - The parameter applies uniformly across all of the plans as all plans must fit within the total of 21 days. The parameter does not have discriminative power. Hence, it should be removed.

3. direct_flights - The parameter does not apply uniformly across all of the plans as different plans can use different direct flights. The parameter has discriminative power. Hence, it should not be removed.

4. city_stays - The parameter does not apply uniformly across all of the plans as different plans can have different city stays. The parameter has discriminative power. Hence, it should not be removed.

5. specific_days - The parameter does not apply uniformly across all of the plans as different plans can meet or not meet specific day requirements. The parameter has discriminative power. Hence, it should not be removed.
<end_of_COT>

<start_of_structured_output>
{
    "combinations": {
        "cities": [
            "Edinburgh",
            "Frankfurt",
            "Dubrovnik",
            "Vienna",
            "London",
            "Budapest",
            "Prague",
            "Valencia",
            "Munich",
            "Oslo"
        ],
        "direct_flights": [
            [
                "Valencia",
                "Munich"
            ],
            [
                "Vienna",
                "Munich"
            ],
            [
                "Vienna",
                "Valencia"
            ],
            [
                "London",
                "Budapest"
            ],
            [
                "London",
                "Oslo"
            ],
            [
                "Edinburgh",
                "Budapest"
            ],
            [
                "Frankfurt",
                "Budapest"
            ],
            [
                "Frankfurt",
                "London"
            ],
            [
                "Prague",
                "Oslo"
            ],
            [
                "Edinburgh",
                "Oslo"
            ],
            [
                "Edinburgh",
                "Munich"
            ],
            [
                "Prague",
                "Munich"
            ],
            [
                "London",
                "Prague"
            ],
            [
                "Edinburgh",
                "London"
            ],
            [
                "Edinburgh",
                "Frankfurt"
            ],
            [
                "Dubrovnik",
                "Munich"
            ],
            [
                "Dubrovnik",
                "Vienna"
            ],
            [
                "Munich",
                "Oslo"
            ],
            [
                "Dubrovnik",
                "Oslo"
            ],
            [
                "Budapest",
                "Munich"
            ],
            [
                "Frankfurt",
                "Prague"
            ],
            [
                "Vienna",
                "London"
            ],
            [
                "Frankfurt",
                "Vienna"
            ],
            [
                "Frankfurt",
                "Oslo"
            ],
            [
                "Frankfurt",
                "Munich"
            ],
            [
                "Vienna",
                "Oslo"
            ],
            [
                "Vienna",
                "Prague"
            ],
            [
                "Budapest",
                "Oslo"
            ],
            [
                "Budapest",
                "Prague"
            ],
            [
                "London",
                "Valencia"
            ],
            [
                "London",
                "Munich"
            ],
            [
                "Frankfurt",
                "Dubrovnik"
            ],
            [
                "Prague",
                "Valencia"
            ],
            [
                "Frankfurt",
                "Valencia"
            ],
            [
                "Edinburgh",
                "Prague"
            ],
            [
                "Vienna",
                "Budapest"
            ]
        ]
    },
    "constraints": {
        "city_stays": {
            "Edinburgh": 5,
            "Frankfurt": 5,
            "Dubrovnik": 2,
            "Vienna": 3,
            "London": 2,
            "Budapest": 3,
            "Prague": 3,
            "Valencia": 2,
            "Munich": 3,
            "Oslo": 2
        },
        "specific_days": {
            "Edinburgh": [
                1,
                5
            ],
            "Oslo": [
                20,
                21
            ],
            "Valencia": [
                17,
                18
            ],
            "London": [
                12,
                13
            ],
            "Munich": [
                18,
                20
            ]
        }
    },
    "combinations_description": "The 'combinations' key contains the parameters for generating possible travel plans. 'cities' is a list of cities to visit, and 'direct_flights' is a list of pairs of cities that have direct flights between them.",
    "constraints_description": "The 'constraints' key contains the conditions that must be met for a valid travel plan. 'city_stays' specifies the number of days to stay in each city, and 'specific_days' specifies the exact days certain cities must be visited for events or meetings."
}
<end_of_structured_output>

<start_of_planning>
{
    "Input Agent": "Identify the set of cities to visit and the direct flight connections between cities. Also, identify the constraints such as the number of days to stay in each city and specific days to be in certain cities for events or meetings.",
    "Combination Function Generator Agent": "Create a function combinations_func() that takes 'cities' and 'direct_flights' as input. Generate all possible sequences of city visits that fit within the total days and only use direct flights.",
    "Solution Function Generator Agent": "Create a function plan_func() that takes 'city_stays' and 'specific_days' as constraints, along with the list of possible plans from combinations_func(). Filter the plans to find one that satisfies all constraints, ensuring all specific day requirements are met."
}
<end_of_planning>
\end{tcblisting}

\begin{tcblisting}{title=Output of Optimization Agent (Fixes Valid Combinations Mistaken for Constraints), listing only, breakable}
<start_of_COT>
1. City Stays - The 'city_stays' constraint specifies the number of days to stay in each city. This constraint exists for every city in the combination space. Hence, it is a combination parameter.
2. Specific Days - The 'specific_days' constraint specifies exact days for certain cities but not for all cities in the combination space. Hence, it is a constraint.
<end_of_COT>

<start_of_structured_output>
{
    "combinations": {
        "cities": [
            "Edinburgh",
            "Frankfurt",
            "Dubrovnik",
            "Vienna",
            "London",
            "Budapest",
            "Prague",
            "Valencia",
            "Munich",
            "Oslo"
        ],
        "direct_flights": [
            [
                "Valencia",
                "Munich"
            ],
            [
                "Vienna",
                "Munich"
            ],
            [
                "Vienna",
                "Valencia"
            ],
            [
                "London",
                "Budapest"
            ],
            [
                "London",
                "Oslo"
            ],
            [
                "Edinburgh",
                "Budapest"
            ],
            [
                "Frankfurt",
                "Budapest"
            ],
            [
                "Frankfurt",
                "London"
            ],
            [
                "Prague",
                "Oslo"
            ],
            [
                "Edinburgh",
                "Oslo"
            ],
            [
                "Edinburgh",
                "Munich"
            ],
            [
                "Prague",
                "Munich"
            ],
            [
                "London",
                "Prague"
            ],
            [
                "Edinburgh",
                "London"
            ],
            [
                "Edinburgh",
                "Frankfurt"
            ],
            [
                "Dubrovnik",
                "Munich"
            ],
            [
                "Dubrovnik",
                "Vienna"
            ],
            [
                "Munich",
                "Oslo"
            ],
            [
                "Dubrovnik",
                "Oslo"
            ],
            [
                "Budapest",
                "Munich"
            ],
            [
                "Frankfurt",
                "Prague"
            ],
            [
                "Vienna",
                "London"
            ],
            [
                "Frankfurt",
                "Vienna"
            ],
            [
                "Frankfurt",
                "Oslo"
            ],
            [
                "Frankfurt",
                "Munich"
            ],
            [
                "Vienna",
                "Oslo"
            ],
            [
                "Vienna",
                "Prague"
            ],
            [
                "Budapest",
                "Oslo"
            ],
            [
                "Budapest",
                "Prague"
            ],
            [
                "London",
                "Valencia"
            ],
            [
                "London",
                "Munich"
            ],
            [
                "Frankfurt",
                "Dubrovnik"
            ],
            [
                "Prague",
                "Valencia"
            ],
            [
                "Frankfurt",
                "Valencia"
            ],
            [
                "Edinburgh",
                "Prague"
            ],
            [
                "Vienna",
                "Budapest"
            ]
        ],
        "city_stays": {
            "Edinburgh": 5,
            "Frankfurt": 5,
            "Dubrovnik": 2,
            "Vienna": 3,
            "London": 2,
            "Budapest": 3,
            "Prague": 3,
            "Valencia": 2,
            "Munich": 3,
            "Oslo": 2
        }
    },
    "constraints": {
        "specific_days": {
            "Edinburgh": [
                1,
                5
            ],
            "Oslo": [
                20,
                21
            ],
            "Valencia": [
                17,
                18
            ],
            "London": [
                12,
                13
            ],
            "Munich": [
                18,
                20
            ]
        }
    },
    "combinations_description": "The 'combinations' key contains the parameters for generating possible travel plans. 'cities' is a list of cities to visit, 'direct_flights' is a list of pairs of cities that have direct flights between them, and 'city_stays' specifies the number of days to stay in each city.",
    "constraints_description": "The 'constraints' key contains the conditions that must be met for a valid travel plan. 'specific_days' specifies the exact days certain cities must be visited for events or meetings."
}
<end_of_structured_output>

<start_of_planning>
{
    "Input Agent": "Identify the set of cities to visit and the direct flight connections between cities. Also, identify the constraints such as specific days to be in certain cities for events or meetings.",
    "Combination Function Generator Agent": "Create a function combinations_func() that takes 'cities', 'direct_flights', and 'city_stays' as input. Generate all possible sequences of city visits that fit within the total days and only use direct flights.",
    "Solution Function Generator Agent": "Create a function plan_func() that takes 'specific_days' as constraints, along with the list of possible plans from combinations_func(). Filter the plans to find one that satisfies all constraints, ensuring all specific day requirements are met."
}
<end_of_planning>
\end{tcblisting}

\begin{tcblisting}{title=Output of Optimization Agent (Validates and Expands Parameters), listing only, breakable}
<start_of_COT>

1. Cities - The list of cities is a fixed set and does not suggest any additional cities should be included. Hence, it is non-expandable.

2. Direct Flights - If a direct flight exists from City A to City B, it is reasonable to assume a direct flight could exist from City B to City A unless explicitly stated otherwise. Hence, it is expandable.

3. City Stays - The number of days to stay in each city is explicitly defined and does not suggest additional days should be added. Hence, it is non-expandable.

4. Specific Days - The specific days for visiting certain cities are explicitly defined for events or meetings. There is no indication that additional days should be added. Hence, it is non-expandable.

<end_of_COT>

<start_of_structured_output>
{
    "combinations": {
        "cities": [
            "Edinburgh",
            "Frankfurt",
            "Dubrovnik",
            "Vienna",
            "London",
            "Budapest",
            "Prague",
            "Valencia",
            "Munich",
            "Oslo"
        ],
        "direct_flights": [
            [
                "Valencia",
                "Munich"
            ],
            [
                "Munich",
                "Valencia"
            ],
            [
                "Vienna",
                "Munich"
            ],
            [
                "Munich",
                "Vienna"
            ],
            [
                "Vienna",
                "Valencia"
            ],
            [
                "Valencia",
                "Vienna"
            ],
            [
                "London",
                "Budapest"
            ],
            [
                "Budapest",
                "London"
            ],
            [
                "London",
                "Oslo"
            ],
            [
                "Oslo",
                "London"
            ],
            [
                "Edinburgh",
                "Budapest"
            ],
            [
                "Budapest",
                "Edinburgh"
            ],
            [
                "Frankfurt",
                "Budapest"
            ],
            [
                "Budapest",
                "Frankfurt"
            ],
            [
                "Frankfurt",
                "London"
            ],
            [
                "London",
                "Frankfurt"
            ],
            [
                "Prague",
                "Oslo"
            ],
            [
                "Oslo",
                "Prague"
            ],
            [
                "Edinburgh",
                "Oslo"
            ],
            [
                "Oslo",
                "Edinburgh"
            ],
            [
                "Edinburgh",
                "Munich"
            ],
            [
                "Munich",
                "Edinburgh"
            ],
            [
                "Prague",
                "Munich"
            ],
            [
                "Munich",
                "Prague"
            ],
            [
                "London",
                "Prague"
            ],
            [
                "Prague",
                "London"
            ],
            [
                "Edinburgh",
                "London"
            ],
            [
                "London",
                "Edinburgh"
            ],
            [
                "Edinburgh",
                "Frankfurt"
            ],
            [
                "Frankfurt",
                "Edinburgh"
            ],
            [
                "Dubrovnik",
                "Munich"
            ],
            [
                "Munich",
                "Dubrovnik"
            ],
            [
                "Dubrovnik",
                "Vienna"
            ],
            [
                "Vienna",
                "Dubrovnik"
            ],
            [
                "Munich",
                "Oslo"
            ],
            [
                "Oslo",
                "Munich"
            ],
            [
                "Dubrovnik",
                "Oslo"
            ],
            [
                "Oslo",
                "Dubrovnik"
            ],
            [
                "Budapest",
                "Munich"
            ],
            [
                "Munich",
                "Budapest"
            ],
            [
                "Frankfurt",
                "Prague"
            ],
            [
                "Prague",
                "Frankfurt"
            ],
            [
                "Vienna",
                "London"
            ],
            [
                "London",
                "Vienna"
            ],
            [
                "Frankfurt",
                "Vienna"
            ],
            [
                "Vienna",
                "Frankfurt"
            ],
            [
                "Frankfurt",
                "Oslo"
            ],
            [
                "Oslo",
                "Frankfurt"
            ],
            [
                "Frankfurt",
                "Munich"
            ],
            [
                "Munich",
                "Frankfurt"
            ],
            [
                "Vienna",
                "Oslo"
            ],
            [
                "Oslo",
                "Vienna"
            ],
            [
                "Vienna",
                "Prague"
            ],
            [
                "Prague",
                "Vienna"
            ],
            [
                "Budapest",
                "Oslo"
            ],
            [
                "Oslo",
                "Budapest"
            ],
            [
                "Budapest",
                "Prague"
            ],
            [
                "Prague",
                "Budapest"
            ],
            [
                "London",
                "Valencia"
            ],
            [
                "Valencia",
                "London"
            ],
            [
                "London",
                "Munich"
            ],
            [
                "Munich",
                "London"
            ],
            [
                "Frankfurt",
                "Dubrovnik"
            ],
            [
                "Dubrovnik",
                "Frankfurt"
            ],
            [
                "Prague",
                "Valencia"
            ],
            [
                "Valencia",
                "Prague"
            ],
            [
                "Frankfurt",
                "Valencia"
            ],
            [
                "Valencia",
                "Frankfurt"
            ],
            [
                "Edinburgh",
                "Prague"
            ],
            [
                "Prague",
                "Edinburgh"
            ],
            [
                "Vienna",
                "Budapest"
            ],
            [
                "Budapest",
                "Vienna"
            ]
        ],
        "city_stays": {
            "Edinburgh": 5,
            "Frankfurt": 5,
            "Dubrovnik": 2,
            "Vienna": 3,
            "London": 2,
            "Budapest": 3,
            "Prague": 3,
            "Valencia": 2,
            "Munich": 3,
            "Oslo": 2
        }
    },
    "constraints": {
        "specific_days": {
            "Edinburgh": [
                1,
                5
            ],
            "Oslo": [
                20,
                21
            ],
            "Valencia": [
                17,
                18
            ],
            "London": [
                12,
                13
            ],
            "Munich": [
                18,
                20
            ]
        }
    },
    "combinations_description": "The 'combinations' key contains the parameters for generating possible travel plans. 'cities' is a list of cities to visit, 'direct_flights' is a list of pairs of cities that have direct flights between them, and 'city_stays' specifies the number of days to stay in each city.",
    "constraints_description": "The 'constraints' key contains the conditions that must be met for a valid travel plan. 'specific_days' specifies the exact days certain cities must be visited for events or meetings."
}
<end_of_structured_output>

<start_of_planning>
{
    "Input Agent": "Identify the set of cities to visit and the direct flight connections between cities. Also, identify the constraints such as specific days to be in certain cities for events or meetings.",
    "Combination Function Generator Agent": "Create a function combinations_func() that takes 'cities', 'direct_flights', and 'city_stays' as input. Generate all possible sequences of city visits that fit within the total days and only use direct flights, considering both directions for each flight.",
    "Solution Function Generator Agent": "Create a function plan_func() that takes 'specific_days' as constraints, along with the list of possible plans from combinations_func(). Filter the plans to find one that satisfies all constraints, ensuring all specific day requirements are met."
}
<end_of_planning>
\end{tcblisting}

\begin{tcblisting}{title=Output of Solution Agent, listing only, breakable}
<start_of_structured_output>
{
    "solutions": [
        {"city": "Edinburgh", "days": [1, 2, 3, 4, 5]},
        {"city": "Frankfurt", "days": [5, 6, 7, 8, 9]},
        {"city": "Dubrovnik", "days": [9, 10]},
        {"city": "Vienna", "days": [10, 11, 12]},
        {"city": "London", "days": [12, 13]},
        {"city": "Budapest", "days": [13, 14, 15]},
        {"city": "Prague", "days": [15, 16, 17]},
        {"city": "Valencia", "days": [17, 18]},
        {"city": "Munich", "days": [18, 19, 20]},
        {"city": "Oslo", "days": [20, 21]}
    ],
    "solutions_description": "The 'solutions' key contains a list of dictionaries, each representing a city visit during the trip. Each dictionary has two keys: 'city' and 'days'. The 'city' key holds the name of the city being visited, and the 'days' key contains a list of integers representing the days of the trip spent in that city."
}
<end_of_structured_output>
\end{tcblisting}

\begin{tcblisting}{title=Output of Combination Function Generation Agent (Requires Reflection), listing only, breakable}
<start_of_COT>
To solve this problem, we need to generate all possible sequences of city visits that fit within the total days and only use direct flights. Here's the step-by-step thought process:

1. **Input Parsing**: We will extract the list of cities, direct flights, and city stays from the input dictionary.

2. **Graph Representation**: We will represent the direct flights as a graph where each city is a node and each direct flight is an edge. This will help us in determining valid paths.

3. **Permutations and Combinations**: We will use permutations to generate possible sequences of city visits. However, we need to ensure that each sequence is valid in terms of direct flights and total days.

4. **Validation of Sequences**:
   - For each permutation of cities, we will check if there is a direct flight between consecutive cities.
   - We will also calculate the total days required for each sequence by summing up the days required for each city stay.
   - If the total days exceed the allowed days, the sequence is invalid.

5. **Output Format**: For each valid sequence, we will create a list of dictionaries where each dictionary contains the city name and the days spent in that city. The days will be represented as a list of integers.

6. **Return the Result**: Finally, we will return a list of all valid sequences in the specified format.

By following these steps, we ensure that the function generates all possible valid travel plans according to the given constraints.
<end_of_COT>

<start_of_code>
from itertools import permutations

def combinations_func(data):
    cities = data['cities']
    direct_flights = data['direct_flights']
    city_stays = data['city_stays']
    
    # Create a set of direct flights for quick lookup
    flight_set = set(tuple(flight) for flight in direct_flights)
    
    # Calculate total days available
    total_days = sum(city_stays.values())
    
    # List to store all valid plans
    valid_plans = []
    
    # Generate all permutations of cities
    for perm in permutations(cities):
        current_day = 1
        valid = True
        plan = []
        
        for i in range(len(perm)):
            city = perm[i]
            stay_days = city_stays[city]
            
            # Check if the current city can be reached from the previous city
            if i > 0:
                prev_city = perm[i - 1]
                if (prev_city, city) not in flight_set:
                    valid = False
                    break
            
            # Add the city and its stay days to the plan
            plan.append({
                "city": city,
                "days": list(range(current_day, current_day + stay_days))
            })
            
            # Update the current day
            current_day += stay_days
            
            # Check if the total days exceed the allowed days
            if current_day > total_days + 1:
                valid = False
                break
        
        # If the plan is valid, add it to the list of valid plans
        if valid:
            valid_plans.append(plan)
    
    return valid_plans
<end_of_code>
\end{tcblisting}

\begin{tcblisting}{title=Output of Combination Function Reflection Agent, listing only, breakable}
<start_of_COT>
The main difference between the ground-truth plan and the output plan is the overlap of days between consecutive cities. In the ground-truth plan, the last day of a city overlaps with the first day of the next city. For example, Edinburgh ends on day 5, and Frankfurt starts on day 5. In the output plan, there is no overlap; each city starts the day after the previous city ends.

To ensure the function generates plans with the same pattern as the ground-truth plan, the thought process should be adjusted to allow for overlapping days between consecutive cities. This means that when calculating the days for each city, the start day of the next city should be the same as the end day of the previous city.

Additionally, the total days calculation should be adjusted to account for this overlap, ensuring that the total number of days does not exceed the available days.
<end_of_COT>

<start_of_COT_correction>
1. **Input Parsing**: Extract the list of cities, direct flights, and city stays from the input dictionary.

2. **Graph Representation**: Represent the direct flights as a graph where each city is a node and each direct flight is an edge. This will help in determining valid paths.

3. **Permutations and Combinations**: Use permutations to generate possible sequences of city visits. Ensure that each sequence is valid in terms of direct flights and total days.

4. **Validation of Sequences**:
   - For each permutation of cities, check if there is a direct flight between consecutive cities.
   - Calculate the total days required for each sequence by summing up the days required for each city stay, allowing for overlap between consecutive cities.
   - If the total days exceed the allowed days, the sequence is invalid.

5. **Output Format**: For each valid sequence, create a list of dictionaries where each dictionary contains the city name and the days spent in that city. The days will be represented as a list of integers, with overlap between consecutive cities.

6. **Return the Result**: Return a list of all valid sequences in the specified format.
<end_of_COT_correction>

<start_of_code_correction>
from itertools import permutations

def combinations_func(data):
    cities = data['cities']
    direct_flights = data['direct_flights']
    city_stays = data['city_stays']
    
    # Create a set of direct flights for quick lookup
    flight_set = set(tuple(flight) for flight in direct_flights)
    
    # Calculate total days available
    total_days = sum(city_stays.values()) - (len(cities) - 1)
    
    # List to store all valid plans
    valid_plans = []
    
    # Generate all permutations of cities
    for perm in permutations(cities):
        current_day = 1
        valid = True
        plan = []
        
        for i in range(len(perm)):
            city = perm[i]
            stay_days = city_stays[city]
            
            # Check if the current city can be reached from the previous city
            if i > 0:
                prev_city = perm[i - 1]
                if (prev_city, city) not in flight_set:
                    valid = False
                    break
            
            # Add the city and its stay days to the plan
            plan.append({
                "city": city,
                "days": list(range(current_day, current_day + stay_days))
            })
            
            # Update the current day, allowing for overlap
            current_day += stay_days - 1
            
            # Check if the total days exceed the allowed days
            if current_day > total_days + 1:
                valid = False
                break
        
        # If the plan is valid, add it to the list of valid plans
        if valid:
            valid_plans.append(plan)
    
    return valid_plans
<end_of_code_correction>
\end{tcblisting}

\begin{tcblisting}{title=Output of Solution Function Generator Agent, listing only, breakable}
<start_of_COT>

1. Cities - The list of cities is a fixed set and does not suggest any additional cities should be included. Hence, it is non-expandable.

2. Direct Flights - If a direct flight exists from City A to City B, it is reasonable to assume a direct flight could exist from City B to City A unless explicitly stated otherwise. Hence, it is expandable.

3. City Stays - The number of days to stay in each city is explicitly defined and does not suggest additional days should be added. Hence, it is non-expandable.

4. Specific Days - The specific days for visiting certain cities are explicitly defined for events or meetings. There is no indication that additional days should be added. Hence, it is non-expandable.

<end_of_COT>

<start_of_structured_output>
{
    "combinations": {
        "cities": [
            "Edinburgh",
            "Frankfurt",
            "Dubrovnik",
            "Vienna",
            "London",
            "Budapest",
            "Prague",
            "Valencia",
            "Munich",
            "Oslo"
        ],
        "direct_flights": [
            [
                "Valencia",
                "Munich"
            ],
            [
                "Munich",
                "Valencia"
            ],
            [
                "Vienna",
                "Munich"
            ],
            [
                "Munich",
                "Vienna"
            ],
            [
                "Vienna",
                "Valencia"
            ],
            [
                "Valencia",
                "Vienna"
            ],
            [
                "London",
                "Budapest"
            ],
            [
                "Budapest",
                "London"
            ],
            [
                "London",
                "Oslo"
            ],
            [
                "Oslo",
                "London"
            ],
            [
                "Edinburgh",
                "Budapest"
            ],
            [
                "Budapest",
                "Edinburgh"
            ],
            [
                "Frankfurt",
                "Budapest"
            ],
            [
                "Budapest",
                "Frankfurt"
            ],
            [
                "Frankfurt",
                "London"
            ],
            [
                "London",
                "Frankfurt"
            ],
            [
                "Prague",
                "Oslo"
            ],
            [
                "Oslo",
                "Prague"
            ],
            [
                "Edinburgh",
                "Oslo"
            ],
            [
                "Oslo",
                "Edinburgh"
            ],
            [
                "Edinburgh",
                "Munich"
            ],
            [
                "Munich",
                "Edinburgh"
            ],
            [
                "Prague",
                "Munich"
            ],
            [
                "Munich",
                "Prague"
            ],
            [
                "London",
                "Prague"
            ],
            [
                "Prague",
                "London"
            ],
            [
                "Edinburgh",
                "London"
            ],
            [
                "London",
                "Edinburgh"
            ],
            [
                "Edinburgh",
                "Frankfurt"
            ],
            [
                "Frankfurt",
                "Edinburgh"
            ],
            [
                "Dubrovnik",
                "Munich"
            ],
            [
                "Munich",
                "Dubrovnik"
            ],
            [
                "Dubrovnik",
                "Vienna"
            ],
            [
                "Vienna",
                "Dubrovnik"
            ],
            [
                "Munich",
                "Oslo"
            ],
            [
                "Oslo",
                "Munich"
            ],
            [
                "Dubrovnik",
                "Oslo"
            ],
            [
                "Oslo",
                "Dubrovnik"
            ],
            [
                "Budapest",
                "Munich"
            ],
            [
                "Munich",
                "Budapest"
            ],
            [
                "Frankfurt",
                "Prague"
            ],
            [
                "Prague",
                "Frankfurt"
            ],
            [
                "Vienna",
                "London"
            ],
            [
                "London",
                "Vienna"
            ],
            [
                "Frankfurt",
                "Vienna"
            ],
            [
                "Vienna",
                "Frankfurt"
            ],
            [
                "Frankfurt",
                "Oslo"
            ],
            [
                "Oslo",
                "Frankfurt"
            ],
            [
                "Frankfurt",
                "Munich"
            ],
            [
                "Munich",
                "Frankfurt"
            ],
            [
                "Vienna",
                "Oslo"
            ],
            [
                "Oslo",
                "Vienna"
            ],
            [
                "Vienna",
                "Prague"
            ],
            [
                "Prague",
                "Vienna"
            ],
            [
                "Budapest",
                "Oslo"
            ],
            [
                "Oslo",
                "Budapest"
            ],
            [
                "Budapest",
                "Prague"
            ],
            [
                "Prague",
                "Budapest"
            ],
            [
                "London",
                "Valencia"
            ],
            [
                "Valencia",
                "London"
            ],
            [
                "London",
                "Munich"
            ],
            [
                "Munich",
                "London"
            ],
            [
                "Frankfurt",
                "Dubrovnik"
            ],
            [
                "Dubrovnik",
                "Frankfurt"
            ],
            [
                "Prague",
                "Valencia"
            ],
            [
                "Valencia",
                "Prague"
            ],
            [
                "Frankfurt",
                "Valencia"
            ],
            [
                "Valencia",
                "Frankfurt"
            ],
            [
                "Edinburgh",
                "Prague"
            ],
            [
                "Prague",
                "Edinburgh"
            ],
            [
                "Vienna",
                "Budapest"
            ],
            [
                "Budapest",
                "Vienna"
            ]
        ],
        "city_stays": {
            "Edinburgh": 5,
            "Frankfurt": 5,
            "Dubrovnik": 2,
            "Vienna": 3,
            "London": 2,
            "Budapest": 3,
            "Prague": 3,
            "Valencia": 2,
            "Munich": 3,
            "Oslo": 2
        }
    },
    "constraints": {
        "specific_days": {
            "Edinburgh": [
                1,
                5
            ],
            "Oslo": [
                20,
                21
            ],
            "Valencia": [
                17,
                18
            ],
            "London": [
                12,
                13
            ],
            "Munich": [
                18,
                20
            ]
        }
    },
    "combinations_description": "The 'combinations' key contains the parameters for generating possible travel plans. 'cities' is a list of cities to visit, 'direct_flights' is a list of pairs of cities that have direct flights between them, and 'city_stays' specifies the number of days to stay in each city.",
    "constraints_description": "The 'constraints' key contains the conditions that must be met for a valid travel plan. 'specific_days' specifies the exact days certain cities must be visited for events or meetings."
}
<end_of_structured_output>

<start_of_planning>
{
    "Input Agent": "Identify the set of cities to visit and the direct flight connections between cities. Also, identify the constraints such as specific days to be in certain cities for events or meetings.",
    "Combination Function Generator Agent": "Create a function combinations_func() that takes 'cities', 'direct_flights', and 'city_stays' as input. Generate all possible sequences of city visits that fit within the total days and only use direct flights, considering both directions for each flight.",
    "Solution Function Generator Agent": "Create a function plan_func() that takes 'specific_days' as constraints, along with the list of possible plans from combinations_func(). Filter the plans to find one that satisfies all constraints, ensuring all specific day requirements are met."
}
<end_of_planning>
\end{tcblisting}

\begin{tcblisting}{title=Output of Solution Function Generator Agent (Does Not Require Reflection), listing only, breakable}
<start_of_COT>
To solve this problem, we need to filter the list of plans based on the constraints provided. The constraints specify certain days that specific cities must be visited. Our task is to find a plan that satisfies all these constraints.

1. **Understand the Input Format**:
   - Each plan is a list of dictionaries, where each dictionary represents a city visit with the city name and the days spent there.
   - Constraints are given in a dictionary format, where 'specific_days' specifies the exact days certain cities must be visited.

2. **Plan the Function**:
   - Iterate over each plan in the 'data' list.
   - For each plan, check if it satisfies all the constraints specified in 'specific_days'.
   - For each city in 'specific_days', ensure that the city is visited on all the required days in the plan.
   - If a plan satisfies all constraints, return that plan.

3. **Implementation Steps**:
   - Loop through each plan in the data.
   - For each plan, loop through each city in the 'specific_days' constraints.
   - Check if the city is present in the plan and if the days match the required days.
   - If all constraints are satisfied for a plan, return that plan.

4. **Edge Cases**:
   - If no plan satisfies the constraints, the function should handle this gracefully (though the problem does not specify what to do in this case, we assume at least one valid plan exists).

By following these steps, we can ensure that the function correctly filters and returns a plan that meets all the specified constraints.
<end_of_COT>

<start_of_code>
def plan_func(data, constraints):
    specific_days = constraints.get('specific_days', {})
    
    for plan in data:
        all_constraints_met = True
        
        for city, required_days in specific_days.items():
            city_found = False
            
            for visit in plan:
                if visit['city'] == city:
                    city_days = set(visit['days'])
                    if all(day in city_days for day in required_days):
                        city_found = True
                        break
            
            if not city_found:
                all_constraints_met = False
                break
        
        if all_constraints_met:
            return plan
<end_of_code>
\end{tcblisting}

\begin{tcblisting}{title=Output of Deliver Function Generator Agent, listing only, breakable}
<start_of_COT>
1. The few-shot example solution begins with a heading that summarizes the trip plan, mentioning the number of cities and days. The format is: "Here is the trip plan for visiting the X European cities for Y days:".
2. The output then details each city visit and the transitions between cities. Each city visit is described with the format: "**Day X-Y:** Visit [City] for Z days.".
3. Transitions between cities are described with the format: "**Day X:** Fly from [City1] to [City2].".
4. The function needs to iterate through the list of city visits, keeping track of the current day and ensuring that transitions and visits are formatted correctly.
5. The function should handle overlapping days correctly, ensuring that transitions are noted on the correct day.

To design the function:
- Start by calculating the total number of cities and days.
- Initialize a string with the heading.
- Iterate through the list of city visits, formatting each visit and transition according to the example.
- Keep track of the current day and adjust for overlaps in days between cities.
- Append each formatted string to the result.
- Return the complete formatted string as the output.
<end_of_COT>

<start_of_code>
def deliver_func(data):
    num_cities = len(data)
    total_days = max(day for city in data for day in city['days'])
    
    result = f"Here is the trip plan for visiting the {num_cities} European cities for {total_days} days:\n\n"
    
    current_day = 1
    for i, visit in enumerate(data):
        city = visit['city']
        days = visit['days']
        start_day = days[0]
        end_day = days[-1]
        visit_days = end_day - start_day + 1
        
        if start_day == current_day:
            result += f"**Day {start_day}-{end_day}:** Visit {city} for {visit_days} days.\n"
        else:
            result += f"**Day {start_day}:** Fly from {data[i-1]['city']} to {city}.\n"
            result += f"**Day {start_day}-{end_day}:** Visit {city} for {visit_days} days.\n"
        
        current_day = end_day + 1
    
    return result
<end_of_code>
\end{tcblisting}

\clearpage
\clearpage
\onecolumn
\section{Other Tables and Figures}

\begin{table*}[htbp]
\centering
\small
\setlength{\abovecaptionskip}{0.2cm}
\setlength{\belowcaptionskip}{-0.0cm}
\renewcommand{\arraystretch}{0.8}
\begin{adjustbox}{max width=\columnwidth}
\begin{tabular}{p{2.5cm}llcccccccccc}
\toprule
\multirow{2}{*}{MODEL} & \multirow{2}{*}{SETTING} 
& \multicolumn{6}{c}{TravelPlanner} 
& \multicolumn{1}{c}{Trip.} 
& \multicolumn{1}{c}{Meet.} \\
\cmidrule(lr){3-8} \cmidrule(lr){9-9} \cmidrule(lr){10-10} 
& & DR & \multicolumn{2}{c}{CPR} & \multicolumn{2}{c}{HCPR} & SR 
& SR & SR \\
\cmidrule(lr){4-5} \cmidrule(lr){6-7}
& & & Micro & Macro & Micro & Macro & 
& & & \\
\midrule
\multirow{4}{*}{GPT-4o}
& Direct       & 100 & 93.6 & 57.0 & 56.4 & 38.7 & 23.6 & 4.0 & 50.7 \\
& CoT   & 100 & 95.2 & 67.0 & 64.8 & 45.2 & 31.5 & 3.4 & 47.4 \\
& SCOPE       & 100 & \textbf{99.7} & \textbf{97.8} & \textbf{97.6} & \textbf{94.8} & \textbf{93.1} & \textbf{87.1} & \textbf{100} \\
\midrule
\multirow{4}{*}{GPT-o3}
& Direct   & 100 & 96.3 & 71.1 & 65.9 & 66.9 & 66.5 & 77.4 & 70.5 \\
& CoT       & 100 & 98.6 & 89.1 & 83.2 & 80.7 & 80.4 & 77.4 & 89.8 \\
& SCOPE      & 100 & \textbf{99.6} & \textbf{97.5} & \textbf{97.2} & \textbf{94.0} & \textbf{92.2} & \textbf{89.1} & \textbf{99.2}\\
\midrule
\multirow{4}{*}{GPT-5}
& Direct   & 100 & 99.6 & 96.6 & 94.3 & 91.4 & 91.2 & 83.8 & 89.3\\
& CoT       & 100 & 99.5 & 96.6 & 94.3 & 91.8 & 91.5 & 85.5 & 93.2 \\
& SCOPE    & 100 &  \textbf{99.7} & \textbf{98.3} & \textbf{97.8} & \textbf{95.3} & \textbf{93.9} & \textbf{88.7} & \textbf{100}\\
\midrule
\multirow{4}{*}{Gemini-1.5-Pro} 
& Direct   & 100 & \textbf{96.8} & 75.0 & 54.6 & 27.5 & 21.8 & 37.6 & 43.2 \\
& CoT       & 100 & 95.1 & 66.8 & 58.7 & 35.5 & 25.3 & 30.9 & 45.4 \\
& SCOPE       & 97.5 & 95.9 & \textbf{85.6} & \textbf{89.3} & \textbf{79.0} & \textbf{71.6} & \textbf{79.3} & \textbf{90.6} \\
\midrule
\multirow{4}{*}{Gemini-2.5-Pro} 
& Direct   & 100 & 99.5 & 95.7 & 93.8 & 92.5 & 91.7 & 87.8 & 95.2 \\
& CoT       & 100 & 99.6 & 96.9 & 96.3 & 94.8 & 93.8 & 87.9 & 96.4 \\
& SCOPE       & 100 & \textbf{99.7} & \textbf{98.2} & \textbf{98.0} & \textbf{95.6} & \textbf{94.2} & \textbf{90.4} & \textbf{99.6} \\
\bottomrule
\end{tabular}
\end{adjustbox}
\caption{Performance comparison across models and settings in inference mode. Full data for TravelPlanner (1000 data points), Trip Planning (1600 data points) and Meeting Planning (1000 data points).}
\label{tab:inference_table_full_data}
\end{table*}

\begin{table*}[htbp]
\centering
\setlength{\abovecaptionskip}{0.2cm}
\setlength{\belowcaptionskip}{0.0cm}
\renewcommand{\arraystretch}{0.8}
\begin{adjustbox}{max width=\textwidth}
\begin{tabular}{p{2.5cm}lcccccccccccc}
\toprule
\multirow{2}{*}{MODEL} & \multirow{2}{*}{SETTING} 
& \multicolumn{4}{c}{TravelPlanner} 
& \multicolumn{4}{c}{Trip} 
& \multicolumn{4}{c}{Meet.} \\
\cmidrule(lr){3-6} \cmidrule(lr){7-10} \cmidrule(lr){11-14}
& & Input & Output & Cost & Time 
& Input & Output & Cost & Time 
& Input & Output & Cost & Time \\
\midrule

\multirow{6}{*}{GPT-4o}
& Direct & \textbf{1055} & 335 & 0.006 & 10 & 2375 & \textbf{163} & \textbf{0.008} & \textbf{6} & 5255 & \textbf{214} & \textbf{0.015} & \textbf{7} \\
& CoT & 1129 & 448 & 0.007 & 14 & 2450 & 474 & 0.011 & 14 & 5330 & 620 & 0.02 & 17 \\
& ToT & 54928 & 5330 & 0.191 & 102 & 17849 & 5396 & 0.099 & 88 & 18382 & 4427 & 0.09 & 114 \\
& EvoAgent & 94166 & 6325 & 0.299 & 111 & 55372 & 3396 & 0.172 & 79 & 74926 & 2480 & 0.212 & 67 \\
& HTP & 89977 & 4104 & 0.266 & 145 & 7236 & 2117 & 0.039 & 50 & 13297 & 942 & 0.043 & 29 \\
& CPMPY & 8249 & 1438 & 0.035 & 17 & 2473 & 693 & 0.013 & 8 & 5353 & 1279 & 0.026 & 12 \\
& SCOPE & 1155 & \textbf{177} & \textbf{0.005} & \textbf{3} & \textbf{2288} & 340 & 0.009 & 13 & \textbf{2004} & 1436 & 0.019 & 295 \\
\midrule
\multirow{6}{*}{GPT-o3}
& Direct & 8157 & 2575 & 0.037 & 55 & \textbf{2381} & 4507 & 0.041 & \textbf{75} & 5261 & 3868 & 0.041 & 77 \\
& CoT & 8231 & 2945 & 0.04 & 43 & 2456 & 4763 & 0.043 & 94 & 5336 & 4851 & 0.049 & \textbf{56} \\
& ToT & 59503 & 17740 & 0.261 & 274 & 22569 & 54201 & 0.479 & 849 & 19782 & 21704 & 0.213 & 412 \\
& EvoAgent & 95440 & 22928 & 0.374 & 359 & 44698 & 35258 & 0.371 & 564 & 76866 & 36651 & 0.447 & 558 \\
& HTP & 90001 & 14512 & 0.296 & 276 & 7015 & 79385 & 0.649 & 1009 & 13600 & 7032 & 0.083 & 191 \\
& CPMPY & 8248 & 3430 & 0.044 & 30 & 2472 & 8532 & 0.073 & 89 & 5352 & 9430 & 0.086 & 91 \\
& SCOPE & \textbf{1756} & \textbf{1151} & \textbf{0.013} & \textbf{17} & 2560 & \textbf{1550} & \textbf{0.018} & 78 & \textbf{2628} & \textbf{3434} & \textbf{0.033} & 152 \\
\midrule
\multirow{6}{*}{GPT-5}
& Direct & 8157 & 4697 & 0.057 & 115 & \textbf{2381} & \textbf{3589} & \textbf{0.039} & 92 & 5261 & 4808 & 0.055 & 170 \\
& CoT & 8231 & 5254 & 0.063 & 142 & 2456 & 7482 & 0.078 & 167 & 5336 & \textbf{4311} & 0.05 & 136 \\
& ToT & 59453 & 36198 & 0.436 & 620 & 19770 & 50897 & 0.534 & 680 & 19950 & 43100 & 0.456 & 532 \\
& HTP & 89931 & 27082 & 0.383 & 391 & 7001 & 46314 & 0.472 & 600 & 13599 & 16244 & 0.179 & 230 \\
& CPMPY & 8248 & 14424 & 0.155 & 181 & 2472 & 7801 & 0.081 & 77 & 5352 & 9760 & 0.104 & \textbf{114} \\
& SCOPE & \textbf{1923} & \textbf{1589} & \textbf{0.018} & \textbf{26} & 2616 & 5082 & 0.054 & \textbf{75} & \textbf{3110} & 4375 & \textbf{0.048} & 241 \\
\midrule
\multirow{3}{*}{Gemini-2.5-Pro}
& Direct & \textbf{1126} & 299 & 0.004 & \textbf{5} & \textbf{2506} & \textbf{195} & \textbf{0.005} & \textbf{3} & 6011 & \textbf{248} & \textbf{0.01} & \textbf{4} \\
& CoT & 1208 & 639 & 0.008 & 10 & 2594 & 749 & 0.011 & 12 & 6099 & 698 & 0.015 & 11 \\
& SCOPE & 1286 & \textbf{181} & \textbf{0.003} & 7 & 2648 & 369 & 0.007 & 55 & \textbf{2311} & 984 & 0.013 & 105 \\
\midrule
\multirow{3}{*}{Gemini-1.5-Pro}
& Direct & \textbf{1126} & \textbf{871} & \textbf{0.014} & 42 & \textbf{2506} & \textbf{790} & \textbf{0.015} & \textbf{67} & 6011 & \textbf{844} & \textbf{0.02} & \textbf{68} \\
& CoT & 1208 & 2221 & 0.035 & 53 & 2594 & 2316 & 0.038 & 78 & 6099 & 1616 & 0.032 & 74 \\
& SCOPE & 1804 & 1816 & 0.03 & \textbf{16} & 3234 & 6769 & 0.106 & 76 & \textbf{3029} & 6863 & 0.107 & 386 \\
\bottomrule
\end{tabular}
\end{adjustbox}
\caption{Token counts, API cost (USD), and total time (seconds) across models and reasoning strategies for three tasks.}
\label{tab:token-cost-time-full}
\end{table*}
\clearpage
\begin{figure*}[t!]
    \centering
    \setlength{\abovecaptionskip}{0.2cm}
    \setlength{\belowcaptionskip}{-0.3cm}
    \includegraphics[width=1.0\textwidth, height=0.6\textwidth]
    {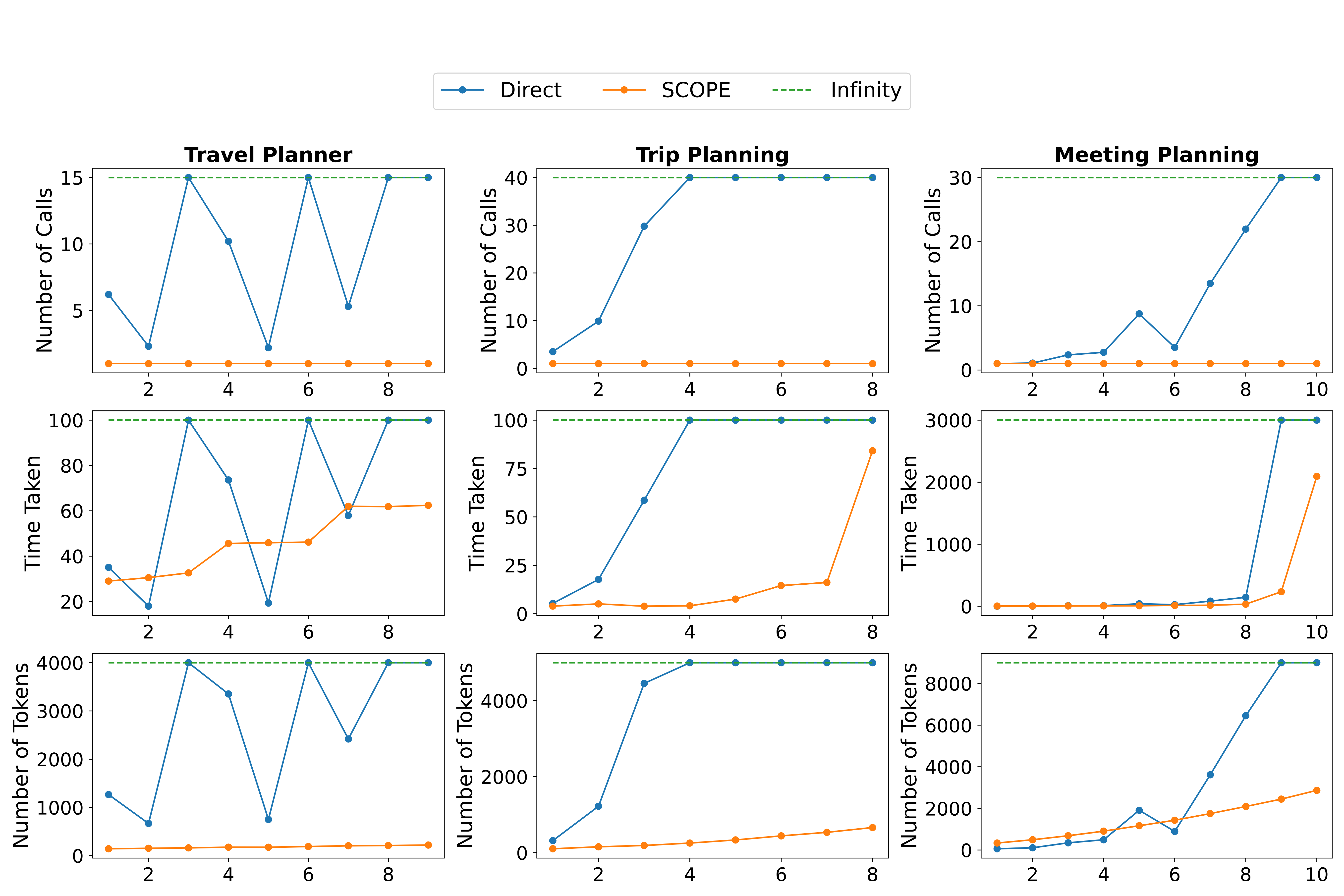}
    \caption{The inference cost required per query to reach correct solution. Top to bottom (Different metrics): Number of calls, time taken, number of tokens. Left to right (Different datasets): Travel Planner, Trip Planning, Meeting Planning. Infinity means an answer cannot be correctly achieved after many repeated attempts.}
    \label{fig:cost_req}
\vspace{-0.3cm}
\end{figure*}

\end{document}